\documentclass[lettersize,journal]{IEEEtran}
\usepackage{amsmath,amsfonts}
\usepackage{algorithmic}
\usepackage{algorithm}
\usepackage{array}
\usepackage[caption=false,font=normalsize,labelfont=sf,textfont=sf]{subfig}
\usepackage{textcomp}
\usepackage{stfloats}
\usepackage{url}
\usepackage{verbatim}
\usepackage{graphicx}
\usepackage{cite}
\usepackage{float}
\usepackage{multirow}
\usepackage{placeins}
\usepackage{listings}
\usepackage{tabularx}
\usepackage{xcolor}

\definecolor{c_comments}{RGB}{34, 139, 34}
\begin{document}

\title{HT-HEDL: High-Throughput Hypothesis Evaluation in Description Logic}

\author{Eyad Algahtani, King Saud University, Saudi Arabia}
        % <-this % stops a space
%\thanks{This paper was produced by the IEEE Publication Technology Group. They are in Piscataway, NJ.}% <-this % stops a space
%\thanks{Manuscript received April 19, 2021; revised August 16, 2021.}}

\markboth{Journal of \LaTeX\ Class Files, December~2024}%
{Shell \MakeLowercase{\textit{et al.}}: A Sample Article Using IEEEtran.cls for IEEE Journals}


\maketitle

\begin{abstract}
We present High-Throughput Hypothesis Evaluation in Description Logic (HT-HEDL). HT-HEDL is a high-performance hypothesis evaluation engine that accelerates hypothesis evaluation computations for inductive logic programming (ILP) learners using description logic (DL) for their knowledge representation; in particular, HT-HEDL targets accelerating computations for the $\mathcal{ALCQI}^{\mathcal{(D)}}$ DL language. HT-HEDL aggregates the computing power of multi-core CPUs with multi-GPUs to improve hypothesis computations at two levels: 1) the evaluation of a single hypothesis and 2) the evaluation of multiple hypotheses (i.e., batch of hypotheses). In the first level, HT-HEDL uses a single GPU or a vectorized multi-threaded CPU to evaluate a single hypothesis. In vectorized multi-threaded CPU evaluation, classical (scalar) CPU multi-threading is combined with CPU's extended vector instructions set to extract more CPU-based performance. The experimental results revealed that HT-HEDL increased performance using CPU-based evaluation (on a single hypothesis): from 20.4 folds using classical multi-threading to $\sim85$ folds using vectorized multi-threading. In the GPU-based evaluation, HT-HEDL achieved speedups of up to $\sim38$ folds for single hypothesis evaluation using a single GPU. To accelerate the evaluation of multiple hypotheses, HT-HEDL combines, in parallel, GPUs with multi-core CPUs to increase evaluation throughput (number of evaluated hypotheses per second). The experimental results revealed that HT-HEDL increased evaluation throughput by up to 29.3 folds using two GPUs and up to $\sim44$ folds using two GPUs combined with a CPU's vectorized multi-threaded evaluation.
\end{abstract}


\section{Introduction}
\label{sec:introduction}
\IEEEPARstart{I}{}nductive logic programming (ILP) is a form of machine learning where background knowledge and learning examples are described using logic-based formalisms. A key advantage of ILP is its ability to learn expressive white-box models from multi-relational data for description and prediction purposes. ILP has been used in various fields, including aerospace \cite{Feng_1991} and biochemistry \cite{Debnath_et_al_1991} to describe complex concepts and relations. ILP is classically used with first-order logic (FOL), especially Horn clauses. However, in recent years, there has been a growth in the use of Description Logic for knowledge representation in ILP applications. 

Description Logic (DL) is a family of logic-based formalisms for representing knowledge. The expressive power of DL is higher than that of propositional logic and lower than that of FOL. In terms of DL applications, DL is used in medicine (to aid medical diagnosis) and in digital libraries (for classification and data retrieval tasks) \cite{dl_book_2007}. However, one of its major applications is in the semantic web, especially the OWL language. OWL (Web Ontology Language) \cite{owl_ref} is a modeling language that semantically represents knowledge as ontologies. There are three OWL variations: OWL Lite, OWL DL, and OWL Full. Each OWL type has its expressive power, i.e., the degree to which it can describe complex concepts; OWL Lite is the least expressive among the three types, and OWL Full is the most expressive type. A major disadvantage in OWL Full (despite its higher expressive power) is its undecidability, and therefore, reasoning tasks are not supported. In practice, OWL DL is commonly used where DL is the underlying logic-based representation, that supports reasoning tasks through DL-based reasoners such as HermiT \cite{hermit}, FaCT++ \cite{fact++}, and Pellet \cite{pellet}. Despite the OWL variations, OWL relies on Resource Description Framework (RDF) \cite{rdf_ref} and RDFS (RDF Schema), which uses a graph data model to describe OWL's knowledge graph.
 
DL is used in ILP because it provides a trade-off between hypothesis expressivity and the needed computing power, which is sufficient for some ILP applications, such as ILP-based recommender systems \cite{Qomariyah_Kazakov_2017}. Regardless of DL-based and FOL-based ILPs, ILP as a machine learning (ML) technique has an inherent scalability limitation, which is learning from large amounts of data. The scalability limitations in DL-based ILPs, as opposed to FOL-based ILPs, are less pronounced because DL has less expressive power than FOL, which translates to less computational complexity. Regardless, ILP's poor scalability limits its potential applications. Therefore, in this work, we aim to increase evaluation performance for DL-based ILP learners by exploiting the computing power of several parallel heterogeneous processors (i.e., CPUs and GPUs). We aggregate the computing power of these heterogeneous processors to increase evaluation performance at the level of single and multiple hypotheses (i.e., hypothesis evaluation throughput). Increasing the evaluation performance for DL-based ILP learners at these two levels is expected to drastically reduce ILP learning times.
 
The article is structured as follows. In Sec.~\ref{sec:Related_work}, we review the parallel and non-parallel approaches to improve ILP learning in general and DL-based ILP learning in particular; however, the focus is on hypothesis evaluation aspects (the scope of this work). In Sec.~\ref{sec:High_performance_hypothesis_evaluation}, we describe our proposed multi-device algorithm for high-throughput hypothesis evaluation in description logic (HT-HEDL). The next sections detail the implementation of HT-HEDL and the experimental results.

\section{Related work}
\label{sec:Related_work}
Two approaches exist for accelerating hypothesis evaluation, parallel and non-parallel approaches. In non-parallel approaches, the aim is to increase evaluation efficiency such as Query Packs \cite{Blockeel_et_al_2000} (to reduce redundant evaluation computations), and sampling techniques \cite{Srinivasan_1999} (for evaluating hypotheses against a smaller but carefully selected subset from learning data). Moreover, some other non-parallel approaches are used to improve ILP performance by reducing the evaluation search space.

In some cases, hypothesis evaluation is a reasoning problem (e.g., in classical ILPs), where reasoning paths are explored for every generated candidate hypothesis; reducing reasoning search space will reflect in improved evaluation performance. Notably, approaches for reducing reasoning search space exist. For example, domain knowledge about the learning problem can be used to constrain candidate hypothesis construction (e.g., using mode declarations in Progol \cite{Muggleton_1995} and Aleph \cite{Srinivasan_2007}), which prunes away areas of the search space that contain invalid hypotheses; other related approaches include proactive knowledge engineering to simplify knowledge representation (through proper choice of predicates) to reduce reasoning complexity. Other non-parallel approaches improve evaluation performance using less expressive logic, such as propositionalization \cite{Paes_et_al_2006}, which reduces the evaluation time and cost of less expressive hypotheses. 

In parallel approaches, the computing power of several parallel processors are aggregated to accelerate hypothesis evaluation. Because logic programming is the cornerstone for ILP, many parallel approaches have been developed to improve the performance of logic programming languages such as Prolog, Datalog, and Answer Set Programming through CPU-based, GPU-based, and Big Data-based approaches \cite{par_logic_prog}. We classify parallel ILP approaches into shared-memory and distributed-memory environments. In a shared-memory environment, researchers in \cite{Meissner_2009,konclude} proposed parallel reasoning for description logic using multi-core CPUs, while other researchers used GPUs for parallel reasoning in DL \cite{gpu_el_reasoner}. Moreover, other researchers have developed approaches for accelerating querying from RDF data stores using GPU-based approaches \cite{Chantrapornchai_Choksuchat_2018,gpu_rdf_opt,gpu_rdf,gpu_sparql}.

Other parallel approaches outsourced hypothesis evaluation to relational database management systems (RDBMSs) to accelerate evaluation for classical ILPs, where the generated hypotheses are represented using SQL queries \cite{Zeng_et_al_2014}. For parallel approaches in distributed-memory environments, several approaches have been developed \cite{Fonseca_et_al_2009}. Researchers in \cite{Srinivasan_Faruquie_Joshi_2010} used the MapReduce framework \cite{Dean_Ghemawat_2008} to accelerate evaluation by exploiting MapReduce's distributed computing capabilities. In addition, other researchers \cite{Konstantopoulos_2007} developed a distributed evaluation approach for Aleph (P-Progol) by dividing data among processors using the MPI framework \cite{OpenMPI_2020}. To accelerate reasoning on OWL ontologies, researchers in \cite{parting_owl} proposed an approach for handling reasoning tasks on large OWL Lite knowledge bases, which involves partitioning the large OWL knowledge base into smaller partitions; where OWL reasoning is performed on each partition in parallel. To accelerate other OWL variations, researchers in \cite{rdf_sq} proposed a hybrid approach that combines parallel and sequential computing to improve reasoning performance on large OWL-RL knowledge bases. To accelerate OWL DL, researchers in \cite{owl_dl_framework} proposed a parallel framework for accelerating the classification task for OWL DL ontologies; They evaluated their proposed framework using existing OWL DL reasoners. In addition to the aforementioned software-based approaches, dedicated hardware (FPGA-based) accelerators were developed to accelerate ILP computations \cite{eyad_fpga,ilp_fpga}. \cite{eyad_fpga} is a dedicated hardware accelerator for embedded systems that uses HT-HEDL's algorithms and knowledge base representation; the HT-HEDL based hardware accelerator achieved a speedup of up to 48.7 fold on a disjunction operation, where the baseline is the sequential performance of Raspberry Pi 4 \cite{raspberry_pi}.

The literature review highlights the following. First, most parallel approaches focus only on targeting the shared-memory environment. Second, parallel CPU-based approaches are limited to scalar processing only, which overlooks potential performance gains that can be achieved by combining the vector instructions of CPUs (especially, multi-core CPUs) with multi-threading; such combination (vectorized multi-threading) will maximize the computing power extracted from CPUs. In the third observation, the use of GPUs for accelerating evaluation is less common (as opposed to CPU approaches); we also observe that using multiple GPUs for evaluation is non-existent. All reviewed approaches (both parallel and non-parallel) focus on improving evaluation at the level of a single hypothesis only, and not directly improving evaluation in terms of evaluation throughput (i.e., increasing the number of evaluated hypotheses per second).
In the next section, we describe HT-HEDL, our proposed evaluation engine. HT-HEDL aims to accelerate hypothesis evaluation at the level of a single hypothesis and evaluation throughput using a heterogeneous combination of multi-GPUs with multi-core CPUs (combined with vector instructions).

\section{High-performance hypothesis evaluation}
\label{sec:High_performance_hypothesis_evaluation}
In this section, we describe HT-HEDL architecture. HT-HEDL is directly based on and builds upon the work in Chapter 5 (the hypothesis evaluation component) of our previous work, SPILDL \cite{Eyad_2020} (a scalable and parallel inductive learner in description logic). Notably, some aspects of SPILDL were used in our previous works \cite{Eyad_Kazakov_2018,Eyad_Kazakov_2019}.
HT-HEDL accelerates hypothesis evaluation for DL-based ILP learners and can evaluate hypotheses up to the $\mathcal{ALCQI}^{\mathcal{(D)}}$ DL language. HT-HEDL accelerates evaluation for a single hypothesis using a single GPU or a single multi-core CPU. HT-HEDL also accelerates evaluation for multiple hypotheses using a combination of heterogeneous processors (multiple GPUs with a multi-core CPU), where a single GPU or a multi-core CPU evaluates a subset of these hypotheses in parallel (with other CPUs and GPUs); this amplifies the performance gains of a single GPU or a CPU (i.e., increased evaluation throughput), which can potentially reduce evaluation times for DL-based ILP learners by an order of magnitude. Next, we describe HT-HEDL's knowledge representation.

\subsection{Knowledge representation}
The original matrix-based representation in Chapter 5 of SPILDL, is designed to facilitate GPU-based computations in general, but optimizing memory access patterns for the original matrix-based representation was not the primary goal. For example, in the original matrix-based representation, the Concepts matrix, which also includes the results matrix, stores all concept memberships of a single individual in continuous memory addresses; in contrast, the memberships of a single concept for all individuals are loaded and processed in batches in the conjunction and disjunction operations. In other words, the original layout of the Concepts matrix (including the results matrix), exactly supports the opposite memory access patterns needed for conjunction and disjunction operations which reduces the computational efficiency of GPUs, even though GPUs still provide good performance speedups. Typical GPUs have sophisticated gather and scatter instructions, which enables them to perform SIMD-based computations on data items that are not stored in continuous memory addresses, as long as a matrix-based representation is used. GPUs are designed and optimized for matrix-based operations. However, poor memory access patterns, even on matrix-based representations, prevent GPUs from reaching their maximum computational performance. Moreover, CPU-based performance in SPILDL is limited to only multithreading on multi-core CPUs because the extended SIMD instruction set available in many modern CPUs requires strict constraints on data alignment to use those SIMD instructions to improve CPU performance. In other words, SIMD capabilities in many modern CPUs are not as sophisticated as SIMD capabilities in GPUs.

\begin{figure}[!htbp]
\centering
\includegraphics[width=0.48\textwidth]{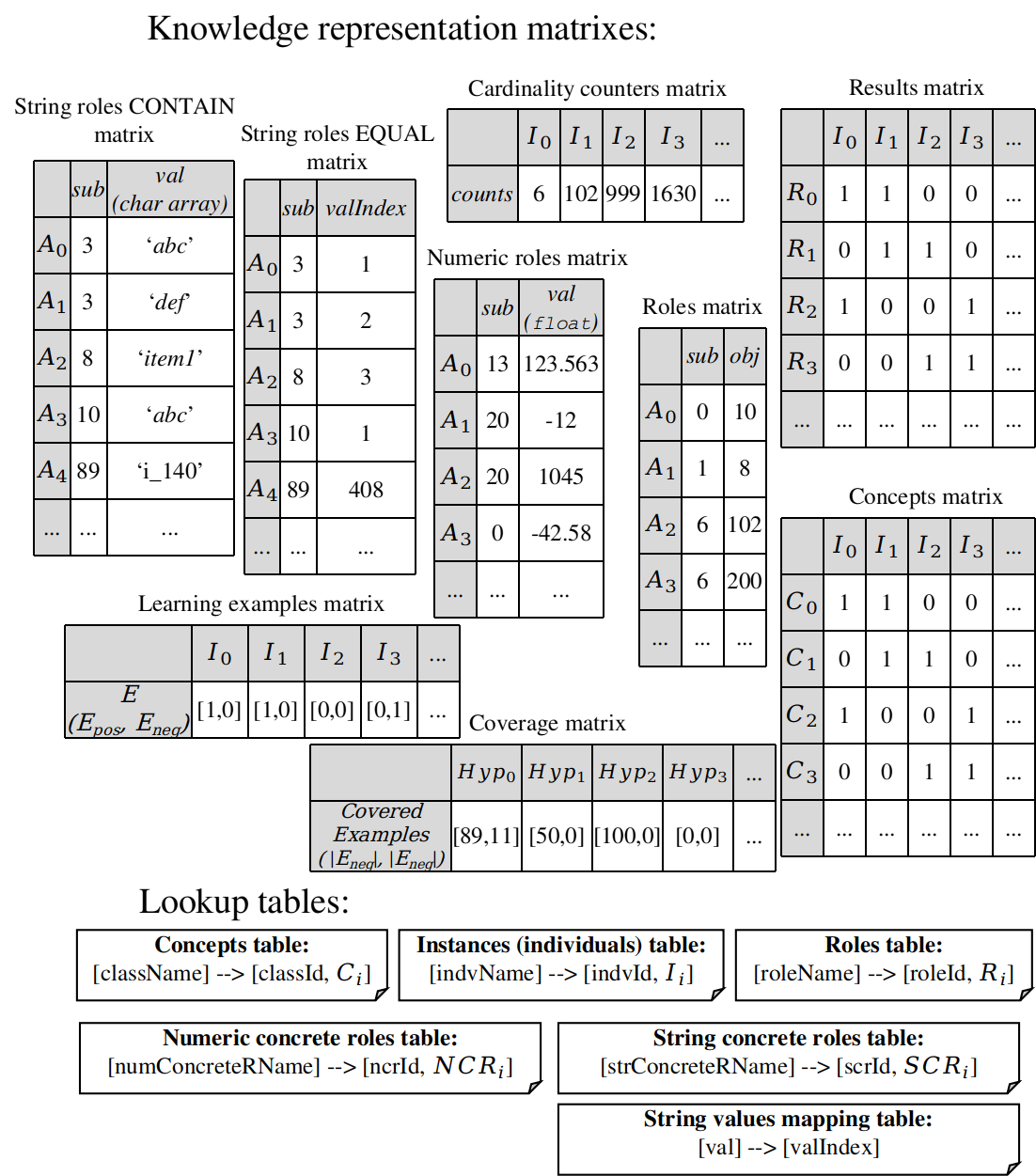}
\caption{HT-HEDL knowledge representation.}
\label{fig:3}    
\end{figure}

HT-HEDL addresses the limitations of SPILDL's original matrix-based representation, by re-engineering the original matrix-based representation to support vectorized (SIMD) CPU-based evaluation and more expressive DL operators by proposing additional DL processing algorithms. Similar to SPILDL's original matrix-based representation, HT-HEDL implements the closed world assumption (CWA) and unique world assumption (UWA) for knowledge representation. The layout of HT-HEDL's matrix-based representation is specifically designed to have efficient DDR memory access patterns in general and to be SIMD-friendly in particular. As a result, due to the improved memory access patterns introduced by HT-HEDL, both CPU-based and GPU-based performance will be higher than that of the original matrix-based representation. HT-HEDL uses a matrix-based representation because matrix-based representations force spatial locality on data items where needed data items are located in adjacent (typically continuous) memory addresses; spatial locality enables the usage of fewer memory operations to load or read the same number of data items, which improves performance due to better memory access patterns. Fig.~\ref{fig:3} depicts HT-HEDL's knowledge representation.

\FloatBarrier

In Fig.~\ref{fig:3}, concepts and roles are represented using matrixes, similar to SPILDL's original matrix-based representation. However, here, we transpose the concepts matrix where individuals (instances) are represented by columns, and concepts are represented by rows; this transposition is performed to improve memory access patterns for both GPUs and CPUs (when computing conjunctions or disjunctions) and to enable the usage of vector instructions in CPU-based evaluation (with multi-threading). In addition, we remove the "Role ID" column from the original roles matrix to rely only on the roles offset lookup table to locate roles' assertions. HT-HEDL represents assertions for roles and concrete roles (both string and numeric) using 2-column matrixes, where the first column is the assertion's subject (individual id), and the second column is the assertion's value. For roles, the assertion value is the assertion's object (individual id); for example, given the role assertion $Person\_a~is\_father\_of~Person\_b$, $Person\_a$ is the assertion's subject, $is\_father\_of$ is the assertion's role, and $Person\_b$ is the assertion's object. For concrete roles, the assertion value is an immediate value: float numbers for numeric roles and string values for string roles. The Learning examples and Coverage matrixes are discussed in Sec.~\ref{sec:fullHyp}.

For storing results for DL operations (e.g., conjunctions), we used the results matrix, where each cell represents a binary membership ('1' for yes, '0' for no) for a particular individual against a particular DL operation. Each row in the results matrix represents memberships on different DL operations, where these rows can store intermediate results when evaluating a hypothesis that uses multiple DL operators. The cardinality counters matrix is used to count cardinalities and apply cardinality operators (i.e., MIN, EQUAL, and MAX).

The knowledge representation matrixes of HT-HEDL and SPILDL's original representations are fully stored in memory. In other words, Both HT-HEDL and SPILDL employ almost zero-copy operations for GPU-based hypothesis evaluation, and completely zero-copy operations for CPU-based evaluation. In GPU-based evaluation, all knowledge representation matrixes (in Fig.~\ref{fig:3}) are created and fully stored in the main memory of each GPU, where only compute operations (GPU kernels in particular) are asynchronously sent for execution through PCI express to each GPU. Once a GPU finishes evaluating its assigned hypotheses, the evaluation results (in the coverage matrix) for that GPU are copied to the host (CPU) memory through a single efficient asynchronous call to $cudaMemcpyAsync(...)$ (a function in Nvidia's CUDA API). A single call to $cudaMemcpyAsync(...)$ asynchronously copies evaluation results for multiple hypotheses in the coverage matrix to the host main memory. The coverage matrix is not made in the host memory, where the GPU can store evaluation results directly in the host memory, because the GPU's memory operations on the host memory have much lower bandwidth, which reduces GPU-based performance, as opposed to GPU's operations on its own memory, which will have much higher bandwidth and thus more performance. Moreover, making GPUs operate solely on host memory, especially when multiple GPUs are considered, results in the congestion of memory operations in the host's memory controller, which consequently reduces the utilization efficiency of GPUs. Therefore, GPU compute operations are performed only on the GPU memory, and the final computation results are asynchronously copied to the host's memory, where the $cudaMemcpyAsync(...)$ function asynchronously copies data from GPU's memory to the host memory in a concurrent (but not parallel) manner, to minimize the congestion on the host memory controller. In terms of CPU-based evaluation, all knowledge representation matrixes are fully stored in the CPU's main memory, and hypothesis evaluation operations are completely zero-copy operations.
In the next section, we describe HT-HEDL's supported DL operators using CPU- and GPU-based approaches.

\subsection{DL operators}
\label{sec:DL_operators}
HT-HEDL supports the same GPU-based baseline operators as SPILDL, such as conjunction, disjunction, existential restriction, and universal restriction, but with some optimizations on restriction operations to improve performance. HT-HEDL builds upon SPILDL's operators to support string (textual) concrete roles through additional DL operators, which are EQUAL and CONTAIN string operators. HT-HEDL can evaluate the supported operators using both GPU- and CPU-based algorithms.

HT-HEDL's CPU-based algorithms rely on the explicit vectorization of compute operations to achieve higher CPU-based performance. Although modern compilers typically have an auto-vectorization feature, which automatically compiles a scalar code into a program that uses vector instructions to improve performance, the scalar code must meet certain compiler constraints, and these constraints may differ between different compilers. Relying on a compiler feature to utilize vector instructions to improve performance in HT-HEDL is not feasible because of the following reasons. First, according to the researchers in \cite{amiri_simd}, implicit (auto) vectorization results in different performances with different compilers, whereas explicit vectorization results in almost the same performance across different compilers. Second, the researchers in \cite{maleki_simd} found that explicit vectorization achieves higher performance than implicit vectorization. Third, the data representation layout and its related memory access patterns also affect auto-vectorization capabilities \cite{feng_simd}; the researchers (in \cite{feng_simd}) also found that auto-vectorization is affected by some compiler optimizations, such as instruction combining optimizations. Therefore, HT-HEDL's CPU-based algorithms use explicit vectorization to ensure consistent and high vector-based CPU performance, regardless of the used compiler. Next, we describe the pseudocodes for HT-HEDL's CPU- and GPU-based algorithms for performing DL operations. We used Nvidia's CUDA semantics to describe the pseudocodes for HT-HEDL's GPU-based DL operations. In every GPU-based pseudocode, $blockIdx.x$ refers to GPU block index, $blockDim.x$ refers to GPU block size (which is the number of GPU threads per a single GPU block), and $threadIdx.x$ refers to GPU thread local ID (within its GPU block). The GPU processes each index ($i$) in a separate GPU thread.

\subsubsection{Conjunction and disjunction operations}
HT-HEDL computes conjunction and disjunction using similar (but not identical) pseudocodes. For computing the two operations using a CPU-based and a GPU-based approach, see Algorithm~\ref{alg:1} and Algorithm~\ref{alg:2}, respectively.

\begin{algorithm}
 \caption{CPU-based conjunction/disjunction with negation}
 \label{alg:1}    
 \begin{lstlisting}[numbers=none]
//INPUT: concept IDs list (conjC) & numbers 
//(conjCNum), negation flags for concept IDs 
//(isCncptNeged), number of individuals (indvsNm)
//OUTPUT: individual memberships covered by
//the DL operation (resultsMatrix)
parallel for(var i=0; i<indvsNm; i+=16)
{//set 16 uint8 values to 1 for conj., 0 for disj.
 vector r = setVectorVal(1);
 for(var j=0; j<conjCNum; j++){
  //load concept memberships for 16 individuals
  vector concept = loadVector(
  conceptsMatrix[conjC[j]*indvsNm+i]);
  vector isNegated = setVectorVal(isCncptNeged[j]);
  //'&'(AND) for conj., '|'(OR) for disj.
  r &= (concept ^ isNegated); //'^' = XOR
 }
 //store conj./disj. result for 16 individuals
 vectorStore(resultsMatrix[i],r);}
\end{lstlisting}
\end{algorithm}
\FloatBarrier

In CPU-based conjunction, a parallel (multi-threaded) for-loop is used on all individuals. In each loop iteration, the result vector $r$ is set to 1 for 16 individuals (assuming 128-bit vector length); moreover, in this loop iteration, a nested loop is used on the conjunction's concepts. For every concept in the conjunction, a vector load operation is used to load the current concept's membership for 16 individuals simultaneously. The loaded memberships are then AND-ed (using vector '\&' operation) with the result vector; the same operation is repeated for each concept in the conjunction. Afterward, the result vector will have the final conjunction results for 16 individuals, which is then stored in the result array using vector store operation. In the CPU-based conjunction, a combination of vector (SIMD) instructions with multi-threading is used to increase performance gains.

 \begin{algorithm}[!htbp]
 \caption{GPU-based conjunction/disjunction with negation -- using CUDA semantics}
 \label{alg:2}    
 \begin{lstlisting}[numbers=none]
//INPUT: concept IDs list (conjC) & numbers 
//(conjCNum), negation flags for concept IDs 
//(isCncptNeged), number of individuals (indvsNm)
//OUTPUT: individual memberships covered by
//the DL operation (resultsMatrix)
var i = blockIdx.x * blockDim.x + threadIdx.x;
if (i<indvsNm){
 var r=1; //=0 for disjunction
 for(var j=0; j<conjCNum; j++){
  var concept = conceptsMatrix[conjC[j]*indvsNm+i];
  var isNegated = isCncptNeged[j];
  //'&'(AND) for conj., '|'(OR) for disj.
  r &= (concept ^ isNegated); //'^' = XOR
 }
 resultsMatrix[i] = r;}
\end{lstlisting}
\end{algorithm}
\FloatBarrier

In GPU-based conjunction, the conjunction is computed for 32 individuals simultaneously in a GPU warp (which has 32 threads). Some GPUs can evaluate multiple warps simultaneously; for example, if a GPU can evaluate four warps simultaneously, then that GPU will evaluate 128 individuals ($4\times32$) simultaneously.

The process for computing disjunction is similar to that for conjunction but with some differences, as outlined in the pseudocodes' comments (in Algorithm~\ref{alg:1} and Algorithm~\ref{alg:2}).

\subsubsection{Role restrictions}
HT-HEDL supports the baseline role restrictions, namely existential, universal, and cardinality restrictions, but with some optimizations. See Algorithm~\ref{alg:3} and Algorithm~\ref{alg:4} for CPU-based and GPU-based pseudocodes for existential restriction. In existential restriction, we look for an assertion per subject where the assertion's object has a concept membership in a particular concept (i.e., the restricting concept). If an assertion is found, the result for the assertion's subject is set to 1. In CPU-based existential restriction, a parallel loop is used to parse all assertions for a given role on roles matrix $rolesMatrix$. If an assertion is found for a subject where its object has a concept membership on the restricting concept $restrictConceptId$, then the result (for the subject) is set to 1 (a match is found). Before finding any matching assertion for any subject, the result for the subject is checked first to see if it is already computed (i.e., a matching assertion was already found). If so, all assertions for that subject are skipped. These checks (and skipping of a subject's assertions) are used to reduce unnecessary memory writes to the same memory addresses by parallel threads, which introduces a performance bottleneck due to the serialized memory write operations. These checks are also used with universal and existential restrictions on concrete roles (both numeric and string). In the GPU-based existential restriction, each GPU thread is processing a single assertion in parallel. If a matching assertion is found and the subject's result is still 0 (no match found), then the result for the subject is set to 1.

\begin{algorithm}
 \caption{CPU-based existential restriction on roles}
 \label{alg:3}    
 \begin{lstlisting}[numbers=none]
//INPUT: role's assertions base address 
//(si=roleAssertsOffset[roleId].startIndex),
//role's assertions end address
//(ei=roleAssertsOffset[roleId].endIndex),
//the restriction's concept ID (restrictConceptId), 
//number of individuals (indvsNm)
//OUTPUT: individual memberships covered by
// the DL operation (resultsMatrix)
memset(resultsMatrix,0,indvsNm);//clear matrix
//each CPU core process fixed chunk of role asserts.
parallel for (var i=si; i<=ei; i++){
 subj = rolesMatrix[i].subj;
 //the result for subject (individual) 
 //is already 1 (true)? skip current subject
 if(resultsMatrix[subj] == 1) {
  //get index of 1st assert. with different subject
  for(; rolesMatrix[i].subj==subj && i<=ei; i++);
   continue; //jump to that assertion index
 }
 if(conceptsMatrix[restrictConceptId*indvsNm+
 rolesMatrix[i].obj]==1 && resultsMatrix[subj]==0) 
  resultsMatrix[subj] = 1; }
\end{lstlisting}
\end{algorithm}
\FloatBarrier

\begin{algorithm}
\caption{GPU-based existential restriction on roles -- using CUDA semantics}
\label{alg:4}    
\begin{lstlisting}[numbers=none]
//INPUT: role's assertions base address
//(si=roleAssertsOffset[roleId].startIndex),
//role's assertions end address
//(ei=roleAssertsOffset[roleId].endIndex),
//the restriction's concept ID (restrictConceptId), 
//number of individuals (indvsNm)
//OUTPUT: individual memberships covered by
//the DL operation (resultsMatrix)
//CPU code: send cmd to clear GPU's resultsMatrix
cudaMemsetAsync(resultsMatrix,0,indvsNm);
//GPU kernel code starts:
var i = si+(blockIdx.x * blockDim.x + threadIdx.x);
if(i<=ei){
 subj=rolesMatrix[i].subj;
 if(conceptsMatrix[restrictConceptId*indvsNm+
 rolesMatrix[i].obj]==1 && resultsMatrix[subj]==0)
  resultsMatrix[subj] = 1;}
\end{lstlisting}
\end{algorithm}
\FloatBarrier

For universal restrictions, see Algorithm~\ref{alg:5} and Algorithm~\ref{alg:6} for CPU-based and GPU-based universal restrictions. In universal restriction, we look for subjects that have all their assertions' objects belonging to only the restricting concept. We also look for individuals that have no assertions at all. To compute this restriction, the result for all individuals is set to 1 to establish the assumption that all individuals are matched until proven otherwise. Similar to existential restriction, all assertions are parsed in parallel. If an assertion is found for a subject that has its object as not being a member of the restricting concept, then the result for that subject is set to 0. For GPU-based universal restriction, a similar approach is used.

\begin{algorithm}
 \caption{CPU-based universal restriction on roles}
 \label{alg:5}    
 \begin{lstlisting}[numbers=none]
//INPUT: role's assertions base address
//(si=roleAssertsOffset[roleId].startIndex),
//role's assertions end address
//(ei=roleAssertsOffset[roleId].endIndex),
//the restriction's concept ID (restrictConceptId), 
//number of individuals (indvsNm)
//OUTPUT: individual memberships covered by
//the DL operation (resultsMatrix)
//set resultsMatrix's cells to 1
memset(resultsMatrix,1,indvsNm);
//each CPU core process fixed chunk of role asserts.
parallel for (var i=si; i<=ei; i++){
 subj = rolesMatrix[i].subj;
 //the result for subject (individual) 
 //is already 0 (false)? skip current subject
 if(resultsMatrix[subj] == 0){
  //get index of 1st assert. with different subject
  for(;rolesMatrix[i].subj==subj && i<=ei;i++);
   continue; //jump to that assertion index
 }
 if(conceptsMatrix[restrictConceptId*indvsNm+
 rolesMatrix[i].obj]==0 && resultsMatrix[subj]==1) 
  resultsMatrix[subj] = 0; }
\end{lstlisting}
\end{algorithm}
\FloatBarrier

\begin{algorithm}
\caption{GPU-based universal restriction on roles -- using CUDA semantics}
\label{alg:6}    
\begin{lstlisting}[numbers=none]
//INPUT: role's assertions base address 
//(si=roleAssertsOffset[roleId].startIndex),
//role's assertions end address
//(ei=roleAssertsOffset[roleId].endIndex),
//the restriction's concept ID (restrictConceptId), 
//number of individuals (indvsNm)
//OUTPUT: individual memberships covered by
//the DL operation (resultsMatrix)
//CPU code: send cmd to set GPU's resultsMatrix
//cells to 1
cudaMemsetAsync(resultsMatrix,1,indvsNm);
//GPU kernel code starts:
var i = si+(blockIdx.x * blockDim.x + threadIdx.x);
if(i<=ei){
 subj=rolesMatrix[i].subj;
 if(conceptsMatrix[restrictConceptId*indvsNm+
 rolesMatrix[i].obj]==0 && resultsMatrix[subj]==1)
  resultsMatrix[subj] = 0;}
\end{lstlisting}
\end{algorithm}
\FloatBarrier

Cardinality restrictions are essentially a variation of existential restriction, that is, instead of finding a matching assertion, we count these matching assertions per subject. The subjects that satisfy the cardinality condition (e.g., MIN, MAX, or EXACTLY) on their number of matched assertions, are considered matching subjects (i.e., their results are set to 1). The CPU-based and GPU-based pseudocodes for computing cardinality restriction can be seen in Algorithm~\ref{alg:7} and Algorithm~\ref{alg:8}.

\begin{algorithm}
\caption{CPU-based cardinality restriction on roles}
\label{alg:7}    
\begin{lstlisting}[numbers=none]
//INPUT: role's assertions base address
//(si=roleAssertsOffset[roleId].startIndex),
//role's assertions end address
//(ei=roleAssertsOffset[roleId].endIndex),
//the restriction's concept ID (restrictConceptId), 
//number of individuals (indvsNm)
//OUTPUT: individual memberships covered by
//the DL operation (resultsMatrix)
//clear resultsMatrix & cardinality counters
memset(resultsMatrix,0,indvsNm);
memset(cardinalityCtrMat,0,indvsNm);
//count matching assertions
parallel for(i=si; i<=ei; i++){
 subj = rolesMatrix[i].subj;
 cCount = 0;
 for(;rolesMatrix[i].subj==subj && i<=ei;i++)
  cCount+=conceptsMatrix[restrictConceptId*indvsNm+
  rolesMatrix[i].obj];

 if(cCount > 0)
  atomicMemoryAdd(cardinalityCtrMat[subj],
  cardinalityCount);}
parallel for(i=0;i<indvsNm;i++){//compute cardinality
 //cardinality counter value
 cVal = cardinalityCtrMat[i]; 
 //use one of three cardinality types:
 //MIN: (cVal>=rVal)
 //EXACTLY: (cVal==rVal),
 //MAX: (cVal>0 && cVal<=rVal) 
 resultMat[i] = (cVal>=rVal);}
\end{lstlisting}
\end{algorithm}
\FloatBarrier

There are three steps of computing the cardinality restriction. First, the result array is cleared (similar to existential restriction). Second, the cardinality counters for the subjects are cleared (in the cardinality matrix $cardinalityCtrMat$). Third, after the matching assertions are counted for all subjects, the cardinality condition (or filter) is applied, and the subjects that satisfy the cardinality condition are set to 1. Because cardinality restrictions have more steps than existential and universal restrictions, they will be slower.
Regarding inverse roles, inverse restrictions can be computed by simply swapping assertions' subjects with objects in both CPU-based and GPU-based pseudocodes for: existential, universal, and cardinality restrictions.

\begin{algorithm}[!htbp]
\caption{GPU-based cardinality restriction on roles -- using CUDA semantics}
\label{alg:8}    
\begin{lstlisting}[numbers=none]
//INPUT: role's assertions base address
//(si=roleAssertsOffset[roleId].startIndex),
//role's assertions end address
//(ei=roleAssertsOffset[roleId].endIndex),
//the restriction's concept ID (restrictConceptId), 
//number of individuals (indvsNm)
//OUTPUT: individual memberships covered by
//the DL operation (resultsMatrix)
//CPU code: send cmd to clear GPU's
//resultsMatrix and cardinalityCtrMat
cudaMemsetAsync(resultsMatrix,0,indvsNm);
cudaMemsetAsync(cardinalityCtrMat,0,indvsNm);
//GPU kernel code starts:
var i = si+(blockIdx.x * blockDim.x + threadIdx.x);
if(i<=ei){
 subj=roleAssertionsMat[i].subj;
 if(conceptsMatrix[restrictConceptId*indvsNm+
 rolesMatrix[i].obj]==1)
  atomicMemoryAdd(cardinalityCtrMat[subj],1);}
//CPU code starts: call GPU kernel 
//'applyCardinalityFilter(...)' to apply cardinality, 
//which uses the same logic as the last
//'parallel for' loop in the CPU-based
//cardinality restriction
applyCardinalityFilter(rVal,resultMat,indvsNm,
[MIN|EXACTLY|MAX]);
\end{lstlisting}
\end{algorithm}
\FloatBarrier

\subsubsection{Concrete role restrictions}
HT-HEDL supports restrictions on numeric and string concrete roles. In the original matrix-based representation, each subject can have at most one concrete role assertion; whereas in HT-HEDL's matrix-based representation, each subject can have multiple concrete role assertions. HT-HEDL's concrete role restrictions are existential restrictions on concrete roles. In these concrete role restrictions, we look for an assertion for a subject where its object (value) matches a certain condition: a cardinality condition for numeric roles and a string pattern for string roles. For computing numeric role restrictions, see Algorithm~\ref{alg:9} and Algorithm~\ref{alg:10} for CPU-based and GPU-based pseudocodes, respectively. 

\begin{algorithm}[!htbp]
 \caption{CPU-based existential restriction on numeric roles}
 \label{alg:9}    
 \begin{lstlisting}[numbers=none]
//INPUT: numeric role's assertions base address 
//(si=numConroleAssertsOffset[roleId].startIndex),
//numeric role's assertions end address
//(ei=numConroleAssertsOffset[roleId].endIndex), 
//the restricting numeric value (rVal), 
//number of individuals (indvsNm)
//OUTPUT: individual memberships covered by
//the DL operation (resultsMatrix)
memset(resultsMatrix,0,indvsNm); //clear matrix
parallel for(i=si; i<=ei; i++){
 subj = nmricRolesMatrix[i].subj;
 //the result for subject (individual) 
 //is already 1 (true)? skip current subject
 if(resultsMatrix[subj] == 1){
  //get index of 1st assert. with different subject
  for(;nmricRolesMatrix[i].subj==subj && i<=ei;i++);
   continue; //jump to that assertion index
 }
 //MIN: (val >= rVal), EXACT: (val == rVal),
 //MAX: (val <= rVal)
 if(nmricRolesMatrix[i].val>=rVal &&
  resultsMatrix[subj]==0)
  resultsMatrix[subj] = 1;}
\end{lstlisting}
\end{algorithm}
\FloatBarrier

 \begin{algorithm}[!htbp]
 \caption{GPU-based existential restriction on numeric roles -- using CUDA semantics}
 \label{alg:10}    
 \begin{lstlisting}[numbers=none]
//INPUT: numeric role's assertions base address 
//(si=numConroleAssertsOffset[roleId].startIndex),
//numeric role's assertions end address
//(ei=numConroleAssertsOffset[roleId].endIndex), 
//the restricting numeric value (rVal), 
//number of individuals (indvsNm)
//OUTPUT: individual memberships covered by
//the DL operation (resultsMatrix)
//CPU code: send cmd to clear GPU's resultsMatrix
cudaMemsetAsync(resultsMatrix,0,indvsNm);
//GPU kernel code starts:
var i = si+(blockIdx.x * blockDim.x + threadIdx.x);
if(i<=ei){
 subj=nmricRolesMatrix[i].subj;
 //MIN: (val >= rVal), EXACT: (val == rVal)
 //MAX: (val <= rVal)
 if(nmricRolesMatrix[i].val >= rVal 
 && resultsMatrix[subj]==0)
  resultsMatrix[subj] = 1;}
\end{lstlisting}
\end{algorithm}
\FloatBarrier

In terms of string restrictions, there are two supported variations. First, there is the EQUAL (EXACT) restriction, where we look for an assertion in which its value matches an exact string (e.g., $val==rVal$). In the second variation (the CONTAIN restriction), we look for an assertion where its value contains a substring (e.g., $strContain(val,rVal)$). The CPU-based and GPU-based pseudocodes for computing EQUAL string restriction can be seen in Algorithm~\ref{alg:11} and Algorithm~\ref{alg:12}, respectively.

\begin{algorithm}
\caption{CPU-based EQUAL restriction on string roles}
\label{alg:11}    
\begin{lstlisting}[numbers=none]
//INPUT: string role's assertions base address
//(si=strConroleAssertsOffset[roleId].startIndex),
//string role's assertions end address
//(ei=strConroleAssertsOffset[roleId].endIndex),
//the restricting string value id (rValInd), 
//number of individuals (indvsNm)
//OUTPUT: individual memberships covered by
//the DL operation (resultsMatrix)
memset(resultsMatrix,0,indvsNm); //clear matrix
parallel for(i=si; i<=ei; i++){
 subj = strConcretRoleMat[i].subj;
 //the result for subject (individual) is 
 //already 1 (true)? skip current subject
 if(resultsMatrix[subj] == 1){
  //get index of 1st assert. with different subject
  for(;strConcretRoleMat[i].subj==subj && i<=ei;i++);
   continue; //jump to that assertion index
 }
 if(strConcretRoleMat[i].valIndex==rValInd 
 && resultsMatrix[subj]==0)
  resultsMatrix[subj] = 1;}
\end{lstlisting}
\end{algorithm}
\FloatBarrier

\begin{algorithm}
\caption{GPU-based EQUAL restriction on string roles -- using CUDA semantics}
\label{alg:12} 
 \begin{lstlisting}[numbers=none]
//INPUT: string role's assertions base address
//(si=strConroleAssertsOffset[roleId].startIndex),
//string role's assertions end address
//(ei=strConroleAssertsOffset[roleId].endIndex),
//the restricting string value id (rValInd), 
//number of individuals (indvsNm)
//OUTPUT: individual memberships covered by
//the DL operation (resultsMatrix)
//CPU code: send cmd to clear GPU's resultsMatrix
cudaMemsetAsync(resultsMatrix,0,indvsNm);
//GPU kernel code starts:
var i = si+(blockIdx.x * blockDim.x + threadIdx.x);
if(i<=ei){
 subj=strConcretRoleMat[i].subj;
 if(strConcretRoleMat[i].valIndex==rValInd 
 && resultsMatrix[subj]==0)
  resultsMatrix[subj] = 1;}
\end{lstlisting}   
\end{algorithm}
\FloatBarrier
In Algorithm~\ref{alg:11} and Algorithm~\ref{alg:12}, a match is found, if there is an assertion where its numeric value mapping $valIndex$ is equal to $rValInd$ (the numeric mapping for the restricting string). To improve efficiency for this restriction, short-circuit evaluation can be used, in which the restricting string ($rVal$) can be checked first (before computing the restriction) if it has a $rValInd$ mapping in the $stringValuesMapping$ lookup table. If it does not, the restriction's computation is skipped and the already cleared result array (using $memset(...)$) is the restriction's result -- this avoids unnecessary computations and improves the restriction's computational efficiency.

For computing CONTAIN restrictions, a similar approach to EQUAL restrictions is used. However, instead of comparing two numeric values ($valIndex==rValInd$), we check if the assertion's value contains a substring $rVal$ through $strContain(val,rVal)$. The CPU-based and GPU-based pseudocodes for this restriction can be seen in Algorithm~\ref{alg:13} and Algorithm~\ref{alg:14}, respectively.

\begin{algorithm}
 \caption{CPU-based CONTAIN restriction on string roles}
 \label{alg:13}   
  \begin{lstlisting}[numbers=none]
//INPUT: string role's assertions base address
//(si=strConroleAssertsOffset[roleId].startIndex),
//string role's assertions end address
//(ei=strConroleAssertsOffset[roleId].endIndex),
//the restricting string value (rVal), 
//number of individuals (indvsNm)
//OUTPUT: individual memberships covered by
//the DL operation (resultsMatrix)
memset(resultsMatrix,0,indvsNm);//clear matrix
parallel for(i=si; i<=ei; i++){
 subj = strConcRolecMat[i].subj;
 //the result for subject (individual) 
 //is already 1 (true)? skip current subject
 if(resultsMatrix[subj] == 1){
  //get index of 1st assert. with different subject
  for(;strConcRolecMat[i].subj==subj && i<=ei;i++);
   continue; //jump to that assertion index
 }
 //strContain(txt,subTxt): return true if 'txt'
 //contains 'subTxt'
 if(strContain(strConcRolecMat[i].val,rVal) 
 && resultsMatrix[subj]==0)
  resultsMatrix[subj] = 1;}
\end{lstlisting} 
\end{algorithm}
\FloatBarrier

\begin{algorithm}
\caption{GPU-based CONTAIN restriction on string roles -- using CUDA semantics}
\label{alg:14} 
 \begin{lstlisting}[numbers=none]
//INPUT: string role's assertions base address
//(si=strConroleAssertsOffset[roleId].startIndex),
//string role's assertions end address
//(ei=strConroleAssertsOffset[roleId].endIndex),
//the restricting string value (rVal), 
//number of individuals (indvsNm)
//OUTPUT: individual memberships covered by
//the DL operation (resultsMatrix)
//CPU code: send cmd clear GPU's resultsMatrix
cudaMemsetAsync(resultsMatrix,0,indvsNm);
//GPU kernel code starts:
var i = si+(blockIdx.x * blockDim.x + threadIdx.x);
if(i<=ei){
 subj=strConcRolecMat[i].subj;
 //strContain(txt,subTxt): return true if 'txt'
 //contains 'subTxt'
 if(strContain(strConcRolecMat[i].val,rVal) 
 && resultsMatrix[subj]==0)
  resultsMatrix[subj] = 1;}
\end{lstlisting}   
\end{algorithm}
\FloatBarrier

\section{Full hypothesis evaluation}
\label{sec:fullHyp}
A DL hypothesis is a series of (potentially) nested DL operations evaluated in a particular order. In HT-HEDL, a hypothesis is represented in a single parent data structure that contains the hypothesis's DL operations stored in consecutive memory addresses (e.g., an array of DL operations). In HT-HEDL, the smallest computational unit is a DL operation, that is, a single DL operation will not be divided into smaller computational units. For example, when computing a conjunction (or a disjunction), all concepts are computed in a single step and will not be divided into a series of smaller conjunctions (or disjunctions) of two concepts at a time.

\begin{figure}[!htbp]
\includegraphics[width=0.48\textwidth]{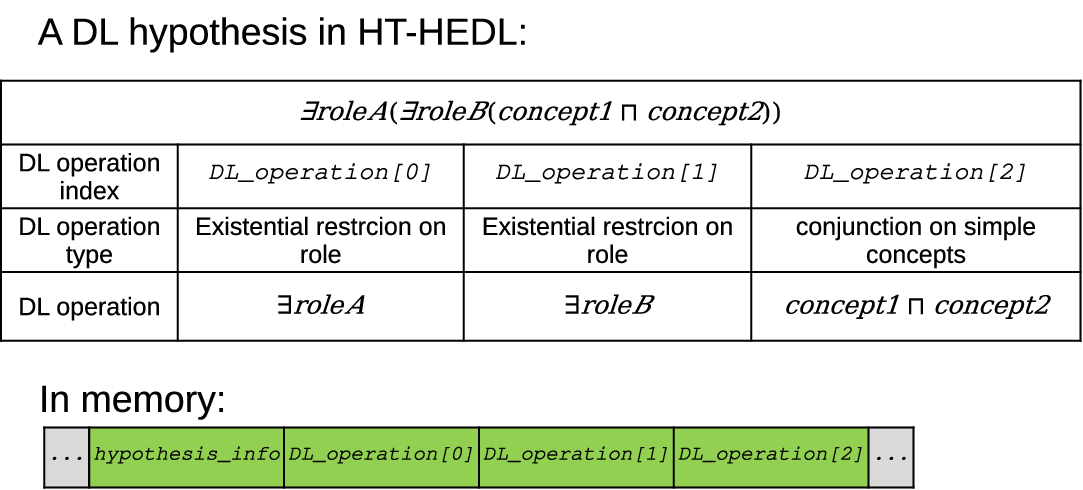}
\caption{DL hypothesis representation in HT-HEDL.}
\label{fig:4}    
\end{figure}

DL hypotheses are inherently tree structures, but storing the hypothesis (including its nodes and DL operations) in consecutive and adjacent memory addresses will improve memory access patterns and thus improve performance because the same hypothesis data can be retrieved using fewer DDR read transactions; this is especially important when processing multiple hypotheses in parallel. See Fig.~\ref{fig:4} for DL hypothesis representation in HT-HEDL. Because a DL hypothesis in HT-HEDL requires fewer DDR memory transactions, memory-bound performance problems are reduced, which makes the CPUs and GPUs perform more computations instead of waiting for data to be retrieved, as opposed to the traditional (less memory efficient) tree-based representations.

\FloatBarrier

To evaluate a hypothesis in HT-HEDL, an evaluation plan is generated. An evaluation plan is essentially a query plan that determines the computational order for DL operations, which also includes the DL operations that are inputs for other DL operations. Once an evaluation plan is generated, it is then executed by either a CPU or a GPU according to the pseudocodes described in Sec.~\ref{sec:DL_operators}. After completing the execution of an evaluation plan, the covered examples are then counted using Algorithm~\ref{alg:15}.

\begin{algorithm}[!htbp]
\caption{Counting covered examples for a single hypothesis}
\label{alg:15} 
 \begin{lstlisting}[numbers=none]
//INPUT: a hypothesis's index (HypIndex), 
//individual memberships covered by the DL operation 
//(resultsMatrix),the positive and negative examples
//from ExMat (learning examples matrix), number of 
//individuals (indvsNm)
//OUTPUT: number of covered positive and negative 
//examples for HypIndex in covExMat (coverage matrix)

//clear covered examples for the hypothesis
covExMat[HypIndex].pos=0;covExMat[HypIndex].neg=0;
parallel reduction for (var i=0; i<indvsNm; i++){
 covExMat[HypIndex].pos+=(resultsMatrix[i] 
 & ExMat[i].pos);
 covExMat[HypIndex].neg+=(resultsMatrix[i] 
 & ExMat[i].neg);}
\end{lstlisting}   
\end{algorithm}
\FloatBarrier

In Algorithm~\ref{alg:15}, we describe a general pseudocode for computing the covered examples. For computing the coverage in CPU-based evaluation, we use multi-threaded parallel reduction for counting covered examples. For GPU-based evaluation, we use the well-known warp-based parallel reduction for counting covered examples; warp-based GPU reductions are known to be the most efficient approach for reduction operations on the GPU.
In this section, we have described how HT-HEDL evaluates a single DL hypothesis. In the next section, we describe how HT-HEDL evaluates multiple hypotheses simultaneously using processors with heterogeneous computing architectures.

\section{Multi-device hypothesis evaluation}
\begin{figure}[!htbp]
\includegraphics[width=0.48\textwidth]{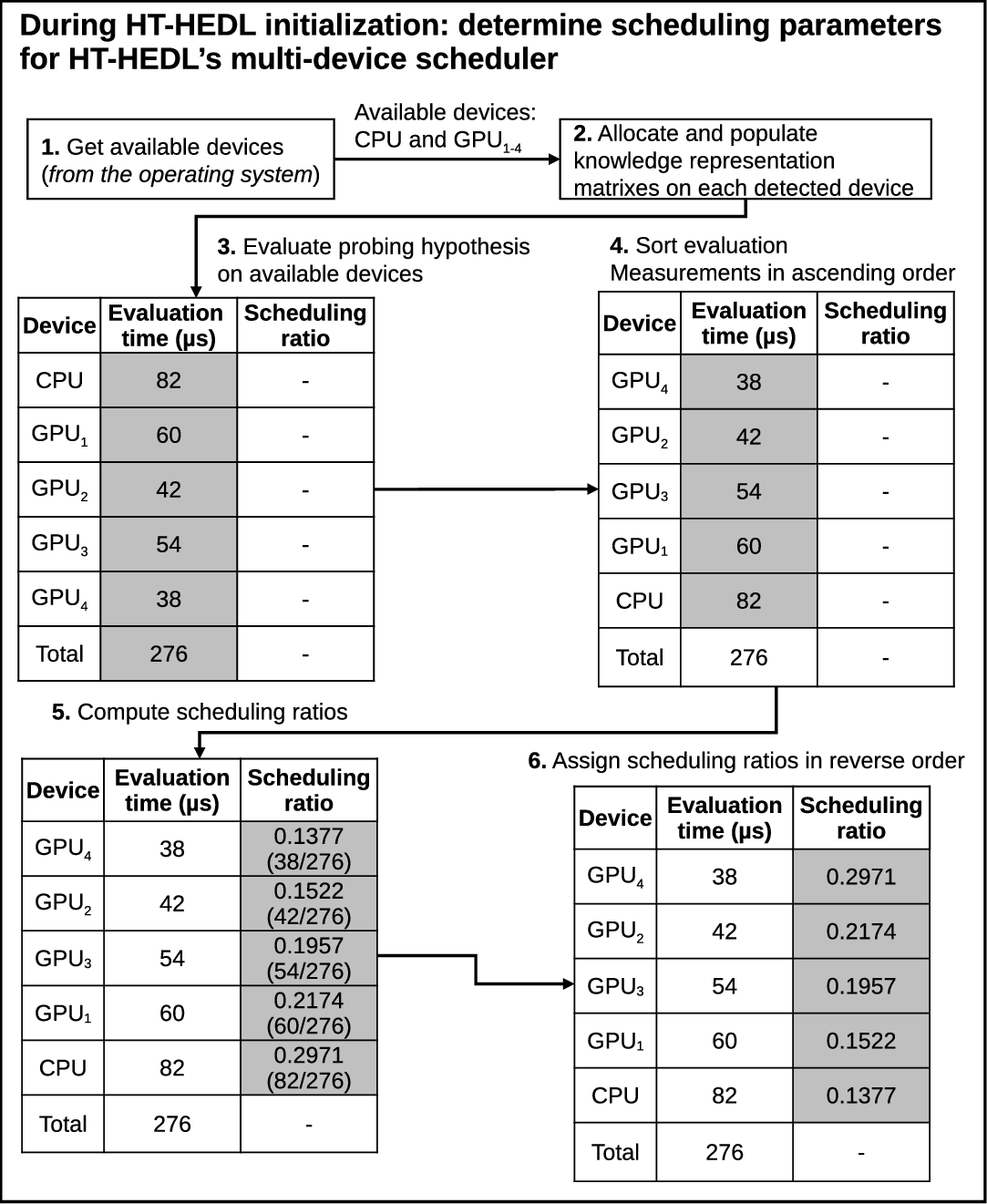}
\caption{Example of HT-HEDL's multi-device scheduling using four GPUs and one multi-core CPU.}
\label{fig:5}    
\end{figure}

In previous sections, we described HT-HEDL's approaches to accelerate the evaluation of a single hypothesis (using a GPU or a multi-core CPU). By combining multiple computing devices (GPUs with CPUs), multiple hypotheses can be evaluated in parallel, which increases the evaluation throughput, thus reducing ILP learning times even more. When evaluating multiple hypotheses in HT-HEDL, HT-HEDL schedules these hypotheses among available evaluation devices (GPUs and CPUs) using a static scheduling strategy, guided by information about each device's evaluation capabilities. To determine the evaluation capabilities for each device, HT-HEDL evaluates a dummy hypothesis (typically a conjunction of concepts) on each device in the system and then measures the device's execution time. The dummy hypothesis is evaluated against the same knowledge base, being used for ILP learning. For example, a knowledge base that has 10 million individuals will result in the dummy hypothesis being evaluated against the same 10 million individuals using every evaluation device in the system. After evaluating the dummy hypothesis on every device against the same knowledge base, HT-HEDL will have an approximation of each device's capability, which is then used to assign the appropriate workload for that device. 
Fig.~\ref{fig:5} presents an example of scheduling evaluation among four GPUs and a single multi-core CPU, to demonstrate HT-HEDL's multi-device scheduling strategy.

\begin{figure}[!htbp]
\includegraphics[width=0.48\textwidth]{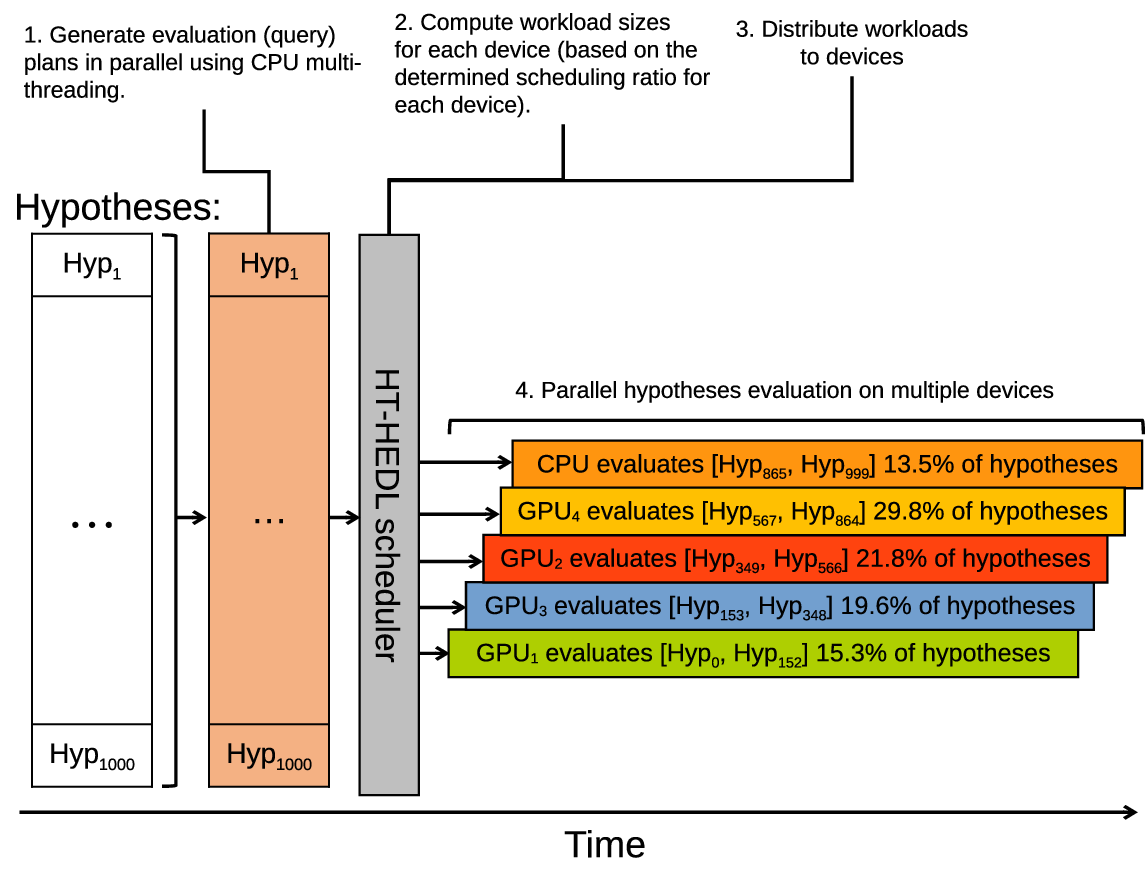}
\caption{Visualization of HT-HEDL's multi-device evaluation of 1000 hypotheses using the same four GPUs and one multi-core CPU.}
\label{fig:6}    
\end{figure}

In Fig.~\ref{fig:5}, during the startup or initialization of HT-HEDL, HT-HEDL first query the operating system for available devices, which results in detecting 4 GPUs and a single multi-core CPU. Second, HT-HEDL allocates and populates knowledge representation matrixes on each detected device, that is, each device will have an exact copy of knowledge representation matrixes. Third, HT-HEDL evaluates the probing hypothesis made of a single conjunction of 5 concepts on each detected device and then measure the device's evaluation time. Executing the same probing hypothesis on every device, will provide information regarding the relative computing capabilities between the detected devices. Once evaluation times are measured, HT-HEDL then computes the scheduling ratio for each device using steps 4-6. These scheduling ratios are computed only once during HT-HEDL's initialization, and then reused to assign a percentage of hypotheses to each device. Even though the total number of hypotheses may vary, yet each device will always evaluate the same ratios of hypotheses. Since HT-HEDL assign workloads to devices based on their evaluation capabilities using relative measures (e.g. ratios), this enables HT-HEDL to maximize multi-device evaluation performance across variable number of hypotheses. HT-HEDL's scheduling algorithm is based on SPILDL's relative load scheduling algorithm; in SPILDL, scheduling ratios and their device assignments are manually provided parameters by the user. Unlike SPILDL, HT-HEDL automatically determines and assigns the scheduling ratios for each multi-core CPU and GPU, based on their computing capabilities -- this is the key novelty in HT-HEDL's scheduling algorithm. Also, SPILDL's scheduling algorithm for hypothesis evaluation only considers GPUs, while HT-HEDL's scheduling algorithm considers both GPUs and multi-core CPUs. Given the nature of HT-HEDL's scheduling algorithm, we can deduce that HT-HEDL uses a form of static scheduling, because HT-HEDL schedules a predetermined (fixed) amount of workloads for each evaluation device. Since HT-HEDL uses static scheduling, it will have minimum (possibly negligible) scheduling overheads in comparison with dynamic scheduling overheads. See Fig~\ref{fig:6} for a demonstration of HT-HEDL's scheduling algorithm using the same hardware (in Fig.~\ref{fig:5}).

\FloatBarrier

As shown in Fig.~\ref{fig:6}, HT-HEDL is assigned to evaluate 1000 hypotheses. Using CPU multithreading, HT-HEDL first generates evaluation plans for the 1000 hypotheses in parallel. Second, HT-HEDL determines the exact number of hypotheses for each evaluation device based on the precomputed scheduling ratios (in Fig.~\ref{fig:5}). Once the exact number of hypotheses is determined for each evaluation device, HT-HEDL then assigns and executes the workload of each device in parallel. Afterward, HT-HEDL waits until parallel compute operations on all devices are completed. In the next section, we describe HT-HEDL's implementation and evaluation details.

\FloatBarrier

\section{Implementation and Evaluation}
\label{sec:imp_eval}
We implemented HT-HEDL in C/C++ and used OpenMP API \cite{openmp} for multithreading. Regarding vector instructions, we used Streaming SIMD Extensions (SSE), a well-known set of vector instructions available in many modern CPUs with x86 and x86-64 (AMD64) architectures; the vector length in SSE is 128-bit long which can process up to 16 (8-bit) concept memberships for 16 individuals simultaneously. There are other implementations for vector instructions (in x86 and x86-64 architectures), such as AVX and AVX-512. However, we chose SSE to ensure compatibility with even older CPUs. Regarding GPGPU, we used Nvidia's CUDA API because it is the most common API in the GPGPU area and also because Nvidia GPUs are used in the experiments.
Regarding evaluation, we used synthetic datasets to evaluate HT-HEDL in best- and worst-case scenarios. The evaluation was conducted using a bottom-up approach, starting from individual DL operators, full hypothesis evaluation, and then multi-device hypothesis evaluation. Table~\ref{tab:1} summarizes HT-HEDL's implementation and machine configuration.

\begin{table}[!htbp]
\caption{HT-HEDL's implementation and machine configuration.}
\label{tab:1}
\begin{tabular}{|l|p{5.2cm}|}
\hline
\multicolumn{2}{|c|}{\textbf{HT-HEDL design and implementation}}\\ \hline
Programming language & C/C++, compiled using Nvidia's $nvcc$ compiler (version: 10.1, V10.1.243) \\ \hline
Multi-threading & OpenMP API\\ \hline
Vector instructions & SSE (using Intel's C/C++ SSE Intrinsics \cite{intel_sse}) \\ \hline
GPGPU & Nvidia CUDA API (using Nvidia CUDA toolkit \cite{cuda_toolkit}, that also include the $nvcc$ compiler) \\ \hline
\multicolumn{2}{|c|}{\textbf{Machine configuration}}\\ \hline
CPU & \shortstack{AMD Ryzen 5950x (16 cores, 32 threads)*}  \\
\hline
Main memory & 32 GB \\
\hline
GPUs & \shortstack{-GPU\textsubscript{1}: Nvidia GTX 1060\\-GPU\textsubscript{2}: Nvidia GTX 970}\\
\hline
\end{tabular}\\
{\raggedright *The CPU was locked at 3.6 GHz on all cores to provide consistent performance.\par}
\end{table}
\FloatBarrier

In the following experimental results, the 'Baseline' column refers to the baseline performance, namely the performance of a single CPU thread without vector instructions. The 'Scalar' column refers to the performance when using all the CPU's native 32 threads without vector instructions, that is, the performance of scalar multi-threading. The 'Vector' column refers to the performance when using all the 32 native threads combined with vector instructions on all cores, that is, the performance of vectorized multi-threading. The 'GPU' column refers to the performance of GPU\textsubscript{2} (Nvidia GTX 970). All time measurements in Table~\ref{tab:2}-Table~\ref{tab:8} are in microseconds and reported using a 5-value average. 
We started the analysis of HT-HEDL by evaluating individual DL operators first. We obtained two sets of experimental results for the conjunction and disjunction operators. In the first experiment, the number of concepts was fixed, but the number of individuals was varied. In the second experiment, the number of concepts was varied, but the number of individuals was fixed. The experimental results for conjunction and disjunction are reported in Table~\ref{tab:2}.

\begin{table}[!htbp]
\caption{Conjunction and disjunction operations on CPU (scalar and vector) and GPU.}
\label{tab:2}
\centering
\begin{tabular}{c|cccc}
\hline
\multicolumn{5}{c}{Conjunction of 5 concepts on \#individuals}\\ \hline
\#individuals & Baseline & Scalar & Vector & GPU\\ \hline
10 & 0 & 0 & 0 & 15\\ \hline
100 & 0 & 3 & 3 & 15\\ \hline
1000 & 5 & 4 & 3 & 16\\ \hline
10,000 & 51 & 7 & 5 & 16\\ \hline
100,000 & 515 & 31 & 12 & 26\\ \hline
1,000,000 & 5168 & 254 & 62 & 137\\ \hline
10,000,000 & 40735 & 2471 & 768 & 1208\\ \hline
\multicolumn{5}{c}{Conjunction of \#concepts on 1 million individuals}\\ \hline
\#concepts & Baseline & Scalar & Vector & GPU\\ \hline
1 & 2741 & 249 & 50 & 63\\ \hline
2 & 3385 & 153 & 50 & 89\\ \hline
4 & 4536 & 227 & 65 & 100\\ \hline
8 & 6699 & 361 & 81 & 144\\ \hline
16 & 10867 & 647 & 143 & 246\\ \hline
32 & 21425 & 1325 & 308 & 441\\ \hline
\multicolumn{5}{c}{Disjunction of 5 concepts on \#individuals}\\ \hline
\#individuals & Baseline & Scalar & Vector & GPU\\ \hline
10 & 0 & 0 & 0 & 15\\ \hline
100 & 0 & 3 & 3 & 15\\ \hline
1000 & 5 & 4 & 3 & 15\\ \hline
10,000 & 50 & 7 & 5 & 16\\ \hline
100,000 & 517 & 30 & 12 & 26\\ \hline
1,000,000 & 5166 & 254 & 64 & 135\\ \hline
10,000,000 & 39970 & 2471 & 728 & 1198\\ \hline
\multicolumn{5}{c}{Disjunction of \#concepts on 1 million individuals}\\ \hline
\#concepts & Baseline & Scalar & Vector & GPU\\ \hline
1 & 2744 & 135 & 55 & 64\\ \hline
2 & 3391 & 152 & 50 & 94\\ \hline
4 & 4561 & 231 & 58 & 99\\ \hline
8 & 6693 & 361 & 86 & 144\\ \hline
16 & 10861 & 649 & 158 & 245\\ \hline
32 & 21351 & 1411 & 346 & 440\\ \hline
\end{tabular}
\end{table}
\FloatBarrier

To evaluate restrictions, we used two (synthetic) datasets. In the first dataset, 'Single subject', we generated role assertions with unique objects but with the same subject. In the second dataset 'Unique subject', we generate role assertions with unique subjects and objects. In the first dataset, parallel memory writes (by role restrictions) were serialized, whereas in the second dataset, memory writes remained in parallel. The performance of role restrictions was highly dependent on the nature of the assertions; therefore, we used these two datasets to provide performance boundaries (i.e., best- and worst-case scenarios) for each restriction. Notably, all reported experiments (including conjunction and disjunction) were conducted using synthetic datasets to provide performance boundaries, and these performance boundaries will reflect HT-HEDL's best-case and worst-case performances against real-world datasets.
Table~\ref{tab:4} presents the experimental results for existential and universal role restrictions. Table~\ref{tab:card_role} reports the cardinality restrictions of the experimental results. Because both MIN and EXACTLY cardinality restrictions have an identical complexity (i.e., a single comparison, '$>=$' for MIN as opposed to '$==$' for EXACTLY), we use the MIN restriction as a representative for both MIN and EXACTLY restrictions.

\begin{table}[!htbp]
\caption{Existential and universal role restrictions using CPU and GPU.}
\label{tab:4}
\begin{tabular}{c|cccccc}
\hline
\multicolumn{7}{c}{Existential restriction}\\ \hline
\multirow{2}{*}{\#assertions} & \multicolumn{3}{c}{Single subject} & \multicolumn{3}{c}{Unique subject}\\
 & Baseline & Scalar & GPU & Baseline & Scalar & GPU\\ \hline
10 & 0 & 0 & 19 & 0 & 0 & 19\\ \hline
100 & 0 & 3 & 19 & 0 & 3 & 19\\ \hline
1,000 & 0 & 3 & 20 & 3 & 3 & 21\\ \hline
10,000 & 5 & 4 & 21 & 26 & 5 & 21\\ \hline
100,000 & 49 & 10 & 34 & 266 & 25 & 30\\ \hline
1,000,000 & 499 & 62 & 139 & 2678 & 168 & 146\\ \hline
10,000,000 & 5259 & 1431 & 1189 & 23477 & 3695 & 1317\\ \hline
\multicolumn{7}{c}{Universal restriction}\\ \hline
\multirow{2}{*}{\#assertions} & \multicolumn{3}{c}{Single subject} & \multicolumn{3}{c}{Unique subject}\\
 & Baseline & Scalar & GPU & Baseline & Scalar & GPU\\ \hline
10 & 0 & 0 & 19 & 0 & 0 & 19\\ \hline
100 & 0 & 3 & 19 & 0 & 3 & 19\\ \hline
1,000 & 0 & 3 & 21 & 2 & 4 & 21\\ \hline
10,000 & 5 & 4 & 22 & 24 & 5 & 21\\ \hline
100,000 & 49 & 9 & 34 & 245 & 22 & 29\\ \hline
1,000,000 & 497 & 61 & 139 & 2446 & 159 & 145\\ \hline
10,000,000 & 5257 & 1461 & 1189 & 22015 & 3708 & 1312\\ \hline

\end{tabular}
\end{table}
\FloatBarrier

\begin{table}[!htbp]
\caption{Cardinality role restriction using CPU and GPU.}
\label{tab:card_role}
\begin{tabular}{c|cccccc}
\hline
\multicolumn{7}{c}{Cardinality restriction (Single subject)}\\ \hline
\multirow{3}{*}{\#assertions} & \multicolumn{6}{c}{Single subject}\\ 
 & \multicolumn{3}{c}{MIN cardinality} & \multicolumn{3}{c}{MAX cardinality}\\
 & Baseline & Scalar & GPU & Baseline & Scalar & GPU\\ \hline
10 & 0 & 8 & 29 & 0 & 7 & 30\\ \hline
100 & 1 & 11 & 30 & 0 & 11 & 30\\ \hline
1,000 & 3 & 11 & 29 & 2 & 11 & 30\\ \hline
10,000 & 26 & 14 & 38 & 22 & 14 & 38\\ \hline
100,000 & 255 & 33 & 135 & 219 & 32 & 135\\ \hline
1,000,000 & 2551 & 212 & 1143 & 2175 & 234 & 1146\\ \hline
10,000,000 & 19320 & 4751 & 10525 & 19879 & 4676 & 10584\\ \hline

\multicolumn{7}{c}{Cardinality restriction (Unique subject)}\\ \hline
\multirow{3}{*}{\#assertions} & \multicolumn{6}{c}{Unique subject}\\ 
 & \multicolumn{3}{c}{MIN cardinality} & \multicolumn{3}{c}{MAX cardinality}\\
 & Baseline & Scalar & GPU & Baseline & Scalar & GPU\\ \hline
10 & 0 & 8 & 29 & 0 & 7 & 28\\ \hline
100 & 1 & 11 & 30 & 1 & 11 & 30\\ \hline
1,000 & 6 & 12 & 30 & 6 & 11 & 30\\ \hline
10,000 & 63 & 28 & 30 & 59 & 28 & 30\\ \hline
100,000 & 642 & 135 & 44 & 605 & 126 & 44\\ \hline
1,000,000 & 6198 & 999 & 241 & 5940 & 982 & 240\\ \hline
10,000,000 & 57228 & 10096 & 2190 & 57590 & 8922 & 2191\\ \hline

\end{tabular}
\end{table}
\FloatBarrier

For concrete role restrictions, we reported experimental results on the 'Single subject' and 'Unique subject' datasets. However, in the context of concrete roles, we generated concrete role assertions for the 'Single subject' and 'Unique subject'. The only difference between these two datasets for roles and concrete roles is in the object of each assertion. For role restrictions, the object is an individual. In concrete role restrictions, the object is an immediate value: numeric for numeric concrete roles, string for string concrete roles. On all generated numeric and string concrete role assertions, the assertion value was fixed to one value. The experimental results for numeric and string concrete role restrictions are reported in Table~\ref{tab:8}. For numeric restrictions, we report experimental results using the MIN cardinality as a representative for MIN, EXACTLY, and MAX cardinalities. Unlike cardinality restrictions on roles, all three cardinality types in numeric concrete role restrictions have identical complexity. For string restrictions, we reported experimental results using EXACTLY and CONTAIN restrictions.

\begin{table}[!htbp]
\caption{Existential restriction on numeric and string concrete roles using CPU and GPU.}
\label{tab:8}
\begin{tabular}{c|cccccc}
\hline
\multicolumn{7}{c}{Numeric concrete roles}\\ \hline
\multirow{2}{*}{\#assertions} & \multicolumn{3}{c}{Single subject} & \multicolumn{3}{c}{Unique subject}\\
 & Baseline & Scalar & GPU & Baseline & Scalar & GPU\\ \hline
10 & 0 & 0 & 19 & 0 & 0 & 19\\ \hline
100 & 0 & 3 & 19 & 0 & 4 & 19\\ \hline
1,000 & 0 & 3 & 20 & 2 & 4 & 20\\ \hline
10,000 & 5 & 4 & 20 & 23 & 5 & 20\\ \hline
100,000 & 49 & 9 & 31 & 238 & 17 & 29\\ \hline
1,000,000 & 497 & 60 & 127 & 2365 & 131 & 135\\ \hline
10,000,000 & 5205 & 1408 & 1042 & 22681 & 1914 & 1184\\ \hline

\multicolumn{7}{c}{String concrete roles (Single subject)}\\ \hline
\multirow{2}{*}{\#assertions} & \multicolumn{3}{c}{EQUAL restriction} & \multicolumn{3}{c}{CONTAIN restriction}\\
 & Baseline & Scalar & GPU & Baseline & Scalar & GPU\\ \hline
10 & 0 & 0 & 19 & 0 & 0 & 23\\ \hline
100 & 0 & 3 & 19 & 0 & 3 & 23\\ \hline
1,000 & 1 & 3 & 20 & 0 & 3 & 28\\ \hline
10,000 & 11 & 4 & 21 & 5 & 6 & 34\\ \hline
100,000 & 113 & 11 & 32 & 57 & 20 & 79\\ \hline
1,000,000 & 1131 & 77 & 126 & 1154 & 214 & 287\\ \hline
10,000,000 & 8340 & 1567 & 957 & 12284 & 7855 & 2346\\ \hline
\multicolumn{7}{c}{String concrete roles (Unique subject)}\\ \hline
\multirow{2}{*}{\#assertions} & \multicolumn{3}{c}{EQUAL restriction} & \multicolumn{3}{c}{CONTAIN restriction}\\
 & Baseline & Scalar & GPU & Baseline & Scalar & GPU\\ \hline
10 & 0 & 0 & 19 & 0 & 0 & 23\\ \hline
100 & 0 & 3 & 19 & 1 & 4 & 24\\ \hline
1,000 & 2 & 4 & 20 & 12 & 4 & 28\\ \hline
10,000 & 23 & 5 & 21 & 128 & 11 & 35\\ \hline
100,000 & 231 & 17 & 29 & 1264 & 72 & 170\\ \hline
1,000,000 & 2314 & 125 & 134 & 11519 & 726 & 1459\\ \hline
10,000,000 & 20916 & 2016 & 1102 & 106916 & 8419 & 12551\\ \hline
\end{tabular}
\end{table}
\FloatBarrier

\begin{table}[!htbp]
\centering
\caption{Summary for evaluating complete hypotheses using CPUs and GPUs.}
\label{tab:11}
\begin{tabularx}{\columnwidth}{c|XXXXXXX}
\hline
\#hyp. & B. & S. & V. & GPU\textsubscript{1} & GPU\textsubscript{2} & GPU\textsubscript{1}, GPU\textsubscript{2} & GPU\textsubscript{1}, GPU\textsubscript{2}, CPU\\ \hline
1 & 4 & 0 & 0 & 0 & 0 & 0* & 0*\\ \hline
10 & 40 & 3 & 1 & 4 & 2 & 1* & 1*\\ \hline
100 & 395 & 31 & 10 & 33 & 25 & 14 & 9\\ \hline
200 & 790 & 62 & 21 & 68 & 50 & 27 & 19\\ \hline
400 & 1579 & 125 & 43 & 134 & 97 & 55 & 38\\ \hline
600 & 2367 & 187 & 64 & 193 & 141 & 83 & 57\\ \hline
800 & 3159 & 250 & 85 & 253 & 185 & 110 & 77\\ \hline
1,000 & 3949 & 312 & 107 & 311 & 231 & 143 & 96\\ \hline
10,000 & 39414 & 3231 & 1079 & 2504 & 2196 & 2255 & 1490\\ \hline
\end{tabularx}\\
{*evaluated by the fastest device (CPU or GPU), based on the probing hypothesis results.}
\end{table}
\FloatBarrier

After evaluating individual DL operators, we evaluated complete DL hypotheses, which included the generation of query/evaluation plans and the computation of DL operators and covered examples; each hypothesis was a conjunctive hypothesis of five concepts against 1 million individuals. We evaluated these hypotheses using single and multiple devices. The experimental results are reported in Table~\ref{tab:11}, where time measurements are in milliseconds and reported using a single reading. In Table~\ref{tab:11}, 'B.' refers to the baseline performance, 'S.' refers to the scalar multi-threaded performance (of the 32 native threads), and 'V.' refers to the vectorized multi-threaded performance. For reference, GPU\textsubscript{1} is 'Nvidia GTX 1060' and GPU\textsubscript{2} is 'Nvidia GTX 970'.

\section{Discussion}
\begin{figure}[!htbp]
\centering
\includegraphics[width=0.48\textwidth]{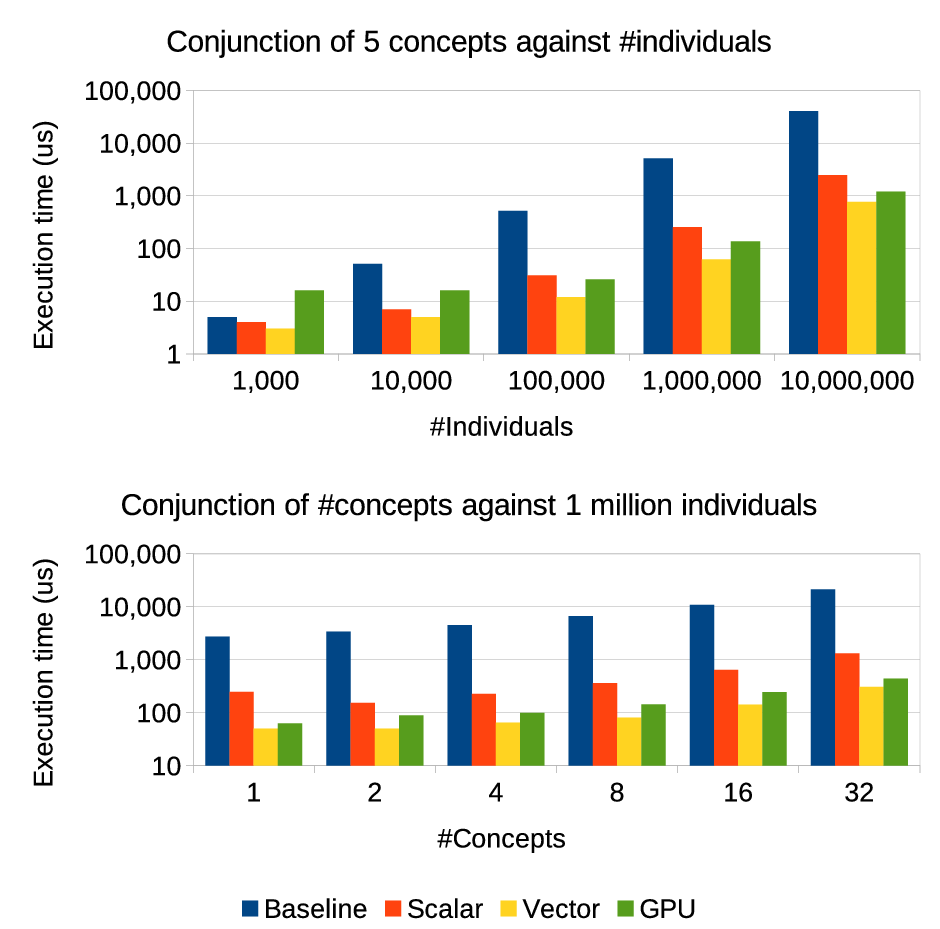}
\caption{Evaluation of conjunction operator.}
\label{fig:7}
\end{figure}

\begin{figure}[!htbp]
\centering
\includegraphics[width=0.48\textwidth]{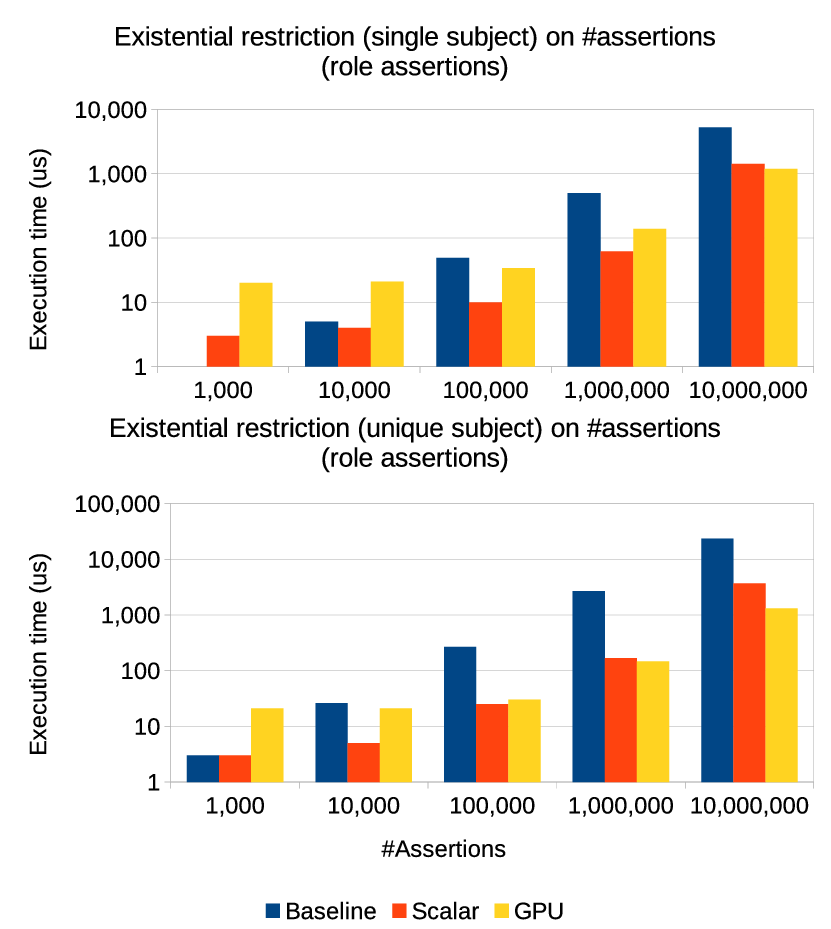}
\caption{The evaluation of an existential restriction operator.}
\label{fig:8}
\end{figure}

According to the experimental results, both CPU- and GPU-based approaches accelerated hypothesis evaluation with varying degrees of speedups. In the conjunction operation, the evaluation of five concepts against one million individuals was accelerated 20.3-fold using scalar multi-threading, 83.4-fold using vectorized multi-threading and 37.7-fold using a GPU. The results for the conjunction operation are summarized in Fig.~\ref{fig:7}; in the disjunction operation, similar speedups were achieved. Regarding role restrictions, the speedups varied between the 'Single subject' and 'Unique subject' datasets; similar to conjunction and disjunction operators, existential and universal restrictions had similar speedups. In the 'Single subject', the GPU achieved lower speedups than in the 'Unique subject' because parallel memory operations were being serialized, which limited the parallel performance of both the CPU and the GPU. In the 'Unique subject' dataset, parallel memory operations were not serialized which enabled more speedups to be achieved. The experimental results for the existential restriction are summarized in Fig.~\ref{fig:8}.

\FloatBarrier

\FloatBarrier

\begin{figure*}[!htbp]
\centering
\includegraphics[width=0.88\textwidth]{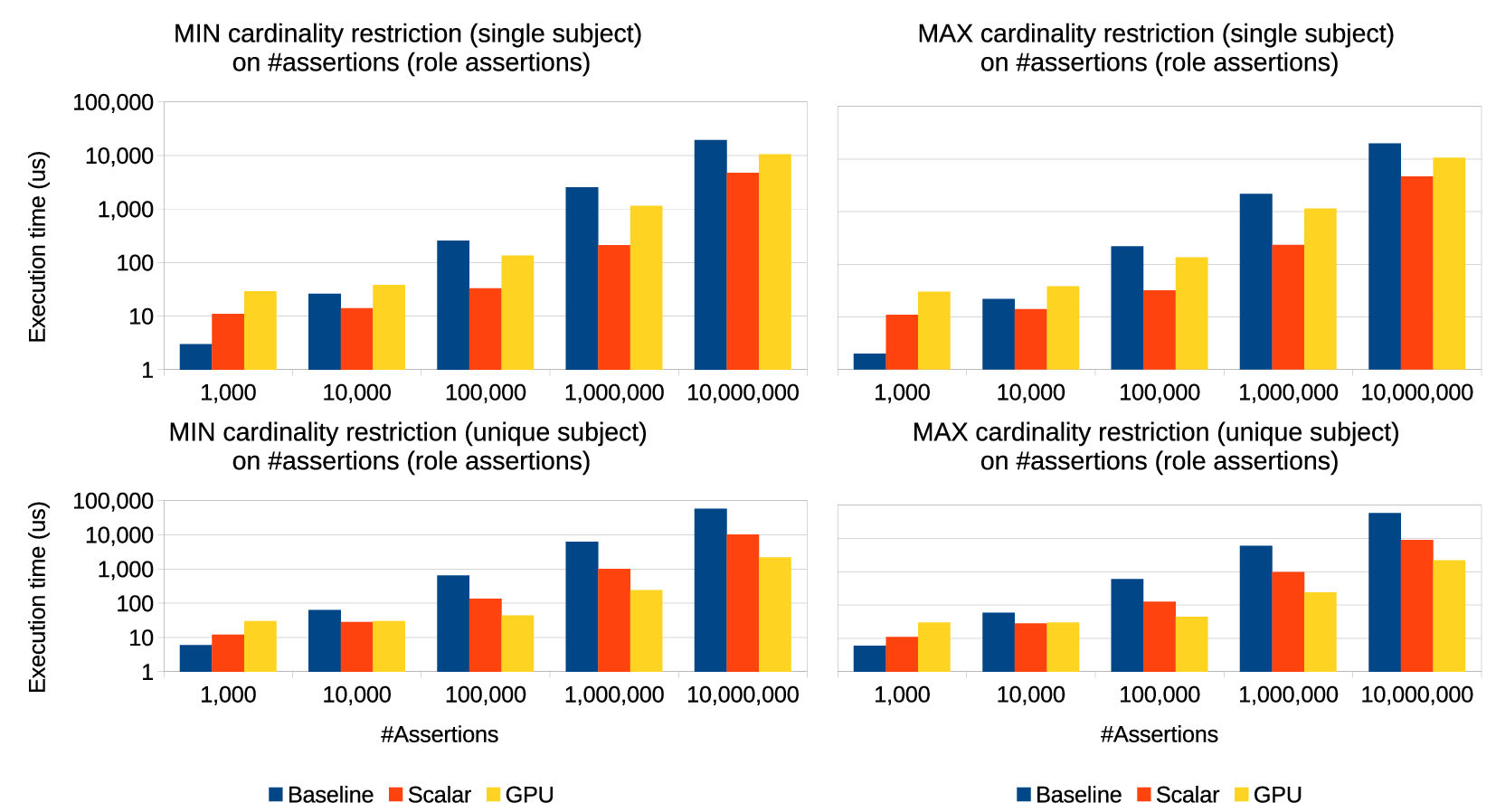}
\caption{Evaluation of cardinality role restriction operator.}
\label{fig:9}
\end{figure*}

Regarding cardinality restrictions, the most number of speedups for 'Single subject' were achieved using a CPU, whereas in 'Unique subject', the GPU achieved the most number of speedups; the sequential performance of GPUs is poorer than that of CPUs. Therefore, serialized memory operations exert a weaker impact on CPU-based performance (as seen in the results for 'Single subject'). Experimental results for cardinality restrictions are summarized in Fig.~\ref{fig:9}.

\begin{figure}[!htbp]
\centering
\includegraphics[width=0.48\textwidth]{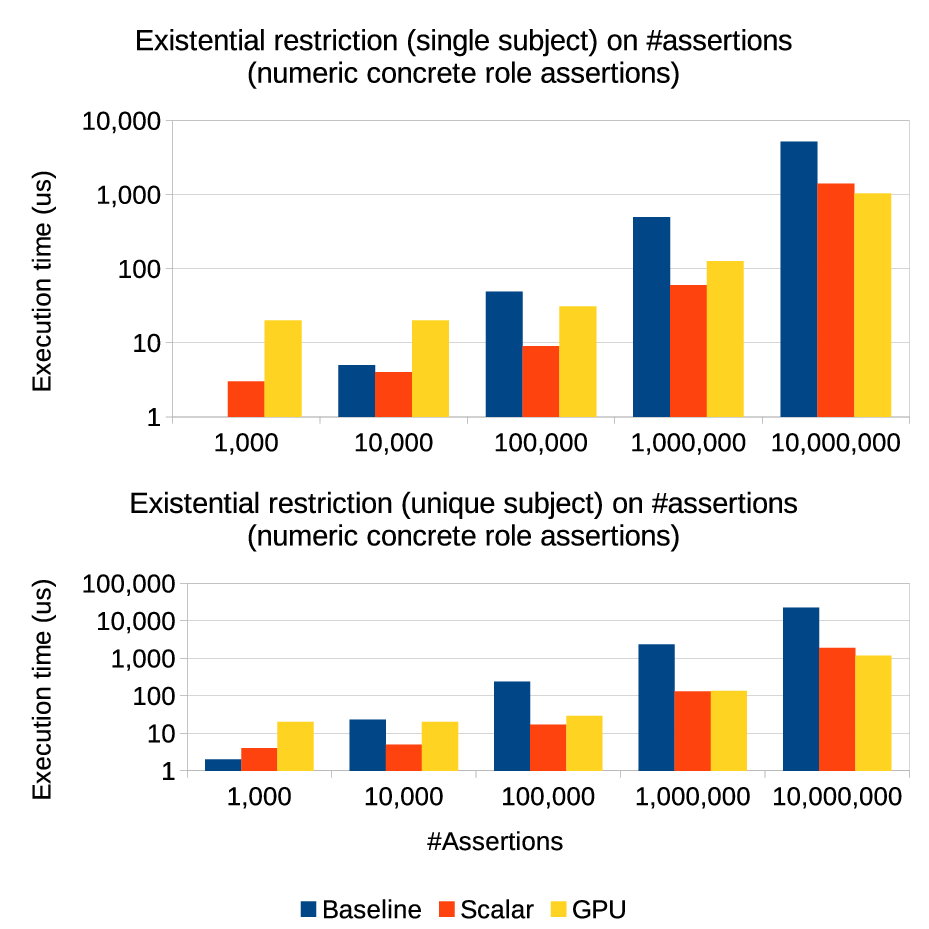}
\caption{The evaluation of numeric concrete role restrictions.}
\label{fig:10}
\end{figure}

\begin{figure*}[!htbp]
\centering
\includegraphics[width=0.88\textwidth]{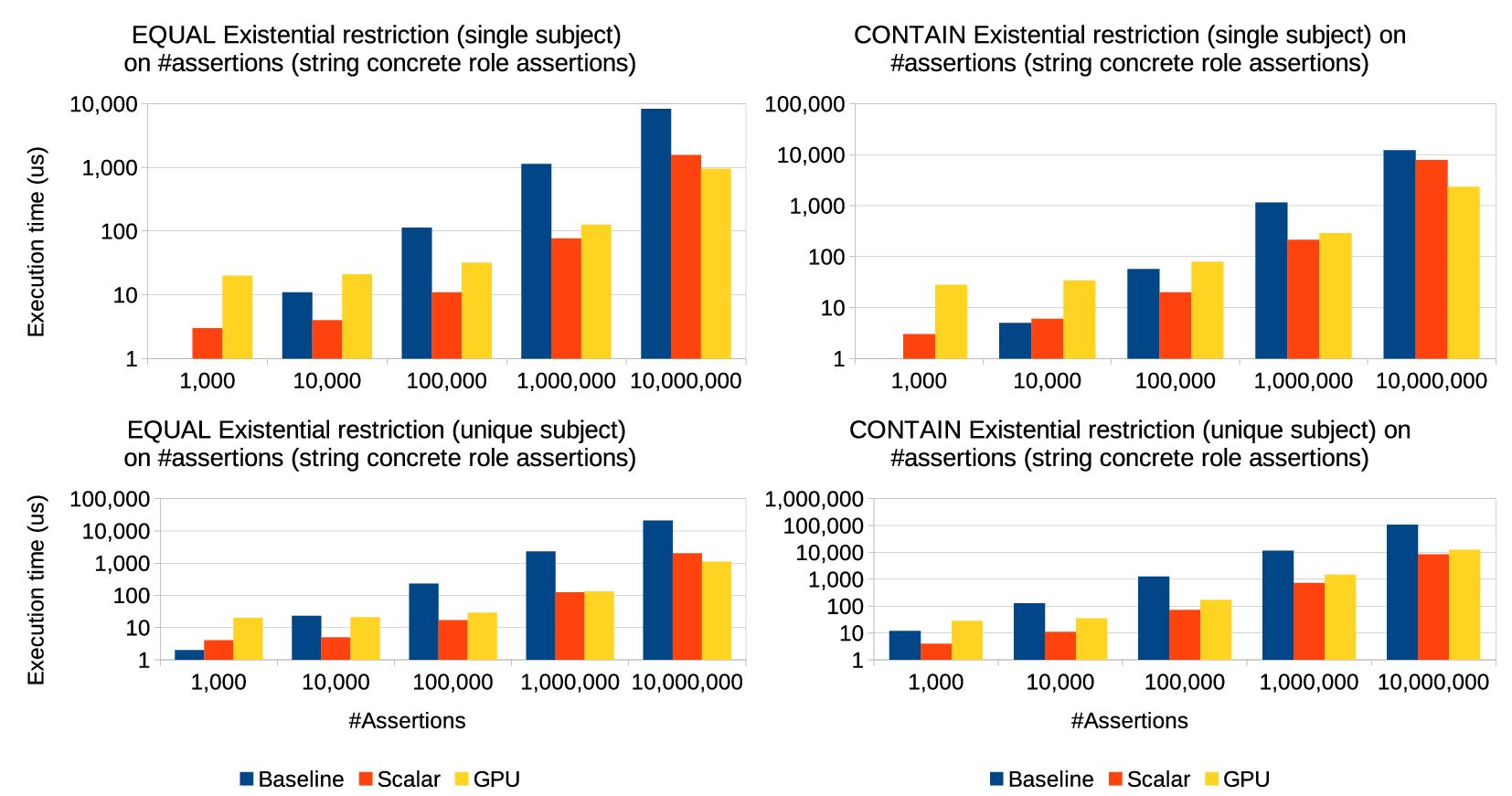}
\caption{The evaluation of string concrete role restrictions.}
\label{fig:11}
\end{figure*}

Regarding concrete role restrictions, the most number of speedups for numeric restrictions were achieved in the 'Unique subject' using both a CPU and a GPU. Similar to role restrictions, most GPU speedups were in the 'Unique subject' dataset. The numeric restriction results are summarized in Fig.~\ref{fig:10}. For string restrictions, both the CPU and the GPU achieved large speedups across the two string restrictions, especially in the 'Unique subject' dataset, whereas the largest speedups were achieved in the EQUAL restriction on the 'Unique subject' dataset. However, the worst-case scenarios for the two string restrictions were observed in the 'Single subject' dataset results on both CPU and GPU results. Fig.~\ref{fig:11} summarizes the string restriction results. The experimental results revealed the CONTAIN restriction is much more suitable for CPU-based evaluation (as opposed to GPU-based evaluation).

\FloatBarrier

\begin{figure}[!htbp]
\centering
\includegraphics[width=0.48\textwidth]{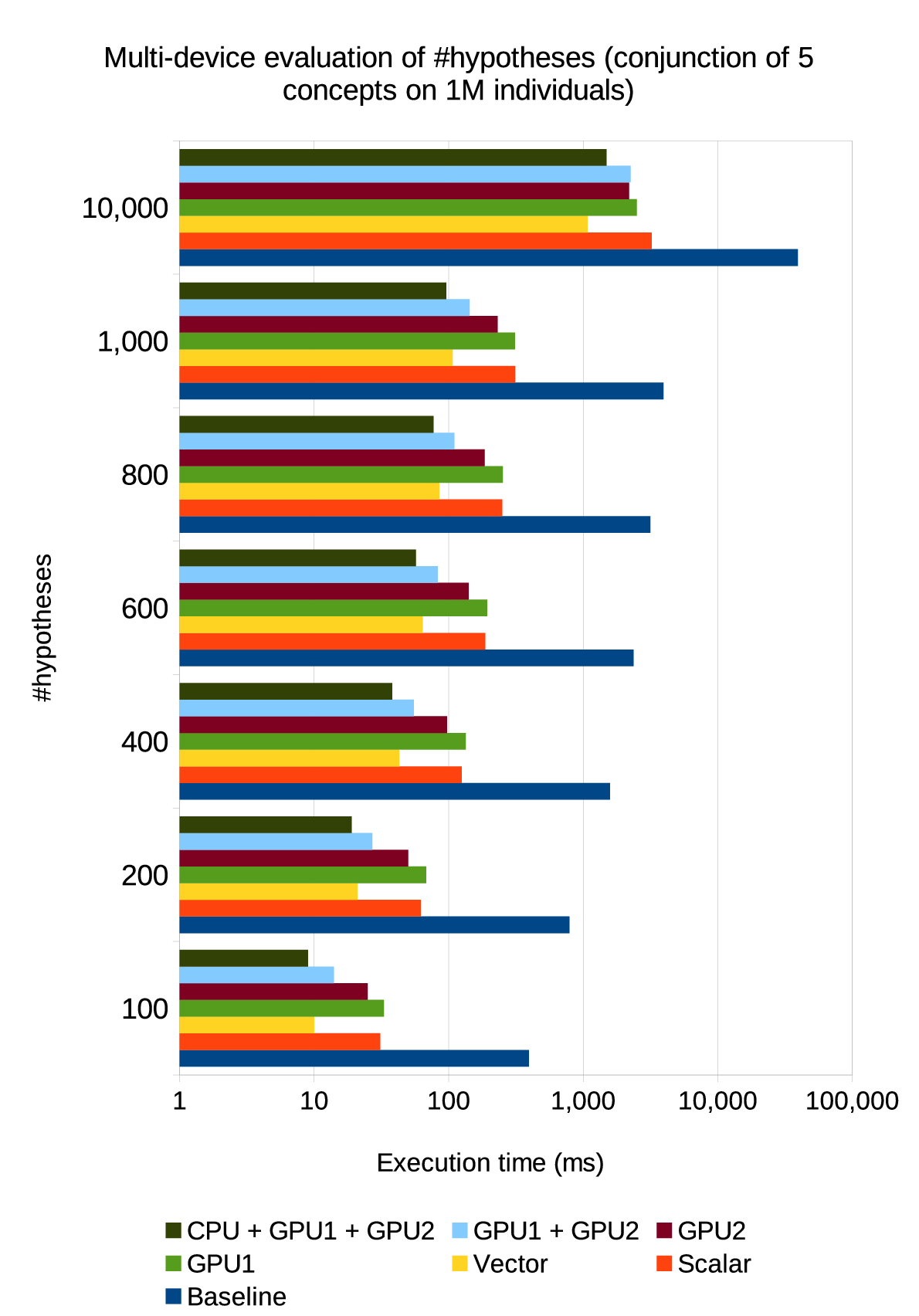}
\caption{Summary for evaluating full hypotheses using single- and multi-device (CPUs + GPUs) approaches.}
\label{fig:12}    
\end{figure}

For evaluating complete hypotheses, the CPU-based (scalar and vector) approaches achieved high speedups, where the largest CPU speedups were achieved using the vectorized approach. Regarding GPU-based evaluation, both GPU\textsubscript{1} (Nvidia GTX 1060) and GPU\textsubscript{2} (Nvidia GTX 970) achieved high speedups in accelerating the evaluation of a single hypothesis. When evaluating 1000 hypotheses, GPU\textsubscript{1} increased performance 12.7-fold, whereas GPU\textsubscript{2} increased the performance 17.1-fold. However, when GPU\textsubscript{1} and GPU\textsubscript{2} were combined, the evaluation performance increased 27.6-fold. In addition, when the CPU was combined with these two GPUs, the evaluation performance increased 41.1-fold. Therefore, adding more processors (a GPU or CPU) increases the hypothesis evaluation throughput. Fig.~\ref{fig:12} summarizes the experimental results for evaluating single and multiple hypotheses using both single and multiple heterogeneous processors.

\FloatBarrier

\FloatBarrier

Fig.~\ref{fig:12} depicts the hypothesis evaluation performance of each GPU and CPU, as well as the combination of GPUs only and CPU+GPUs across different numbers of hypotheses. Noticeably, HT-HEDL's scheduling algorithm performed well when the number of hypotheses was $\leq$ 1000. For a higher number of hypotheses such as 10,000, HT-HEDL's scheduling algorithm performed poorly, as the combination of evaluation devices did not result in higher aggregated performance; however, the aggregated performance was still a major improvement over baseline, which was $\sim17.5$ for GPU\textsubscript{1}+GPU\textsubscript{2}, and $\sim26.5$ for GPU\textsubscript{1}+GPU\textsubscript{2}+CPU. A potential solution to the limitation of HT-HEDL's scheduling algorithm on a large number of hypotheses may be to evaluate the large number of hypotheses (e.g., 10,000 hypotheses) in individual batches of 1000 hypotheses at a time. We chose an individual hypothesis batch of 1000 hypotheses because HT-HEDL's scheduling algorithm stopped providing aggregated performance above 1000 hypotheses. HT-HEDL's scheduling algorithm is not designed to outperform existing scheduling algorithms but rather to aggregate the computing performance of CPUs and GPUs toward improving hypothesis evaluation. Improving and addressing the limitations of HT-HEDL's scheduling algorithm remains a potential direction for future work.

\FloatBarrier

The experimental results for HT-HEDL provide clear evidence that HT-HEDL's CPU- (both scalar and vector) and GPU-based approaches increase the computing performance (drastically) for DL operators. The experimental results also suggest that combining multiple (heterogeneous) processors (e.g., GPUs and CPUs) increases hypothesis evaluation performance by increasing evaluation throughput.
Each acceleration approach either CPU- or GPU-based, has certain advantages and limitations. CPU-based approaches have two key advantages. First, CPUs have virtually zero communication overhead between their cores, whereas GPUs have a small (but not zero) communication overhead (through PCI link) with CPUs. Second, CPU approaches have good memory scalability; that is, more memory can added (as in hardware) to handle larger datasets. On the other hand, GPUs have fixed memory sizes, which limits certain GPUs to certain dataset sizes.
Notably, HT-HEDL's CPU-based approaches are designed to improve performance over baseline. In other words, they should not be expected to be always faster than GPU-based approaches. GPUs typically outperform CPUs, especially on vector processing workloads such as matrix-related math operations. In this work, both scalar and vectorized multithreaded CPU evaluation outperformed the GPU-based evaluation in several experiments because the GPUs used in the experiments (Nvidia GTX 1060 which was released on 2016, and Nvidia GTX 970 which was released on 2014) were several generations older than the CPU that was used (AMD Ryzen 5950x, which was released on 2020). In other words, recent GPUs from the same generation (period) as the used CPU can outperform the CPU, even when using the CPU's vectorized multithreading capability. Specifically, in our other work on MP-HTHEDL \cite{mp_hthedl} (Table 4 in particular), we reported the performance of HT-HEDL on a machine with four heterogeneous GPUs: Nvidia RTX 3090, Nvidia RTX 3070, Nvidia GTX 1070, and Nvidia GTX 970. Despite the heterogeneous mix of old and modern GPUs, which have different computing capabilities, HT-HEDL provided higher aggregated performance using those GPUs. In our other work ( \cite{mp_hthedl}), HT-HEDL achieved the highest aggregated performance using GPUs only. When a CPU+GPUs evaluation was used, HT-HEDL achieved poorer aggregated performance than the GPUs-only evaluation. This is attributable to the speed of modern GPUs (especially the Nvidia RTX 3090), and combining the CPU with these powerful modern GPUs, places the CPU as the performance bottleneck. In a different work on MP-SPILDL \cite{mp_spildl}, we also used HT-HEDL algorithms to improve hypothesis evaluation using both GPUs and CPUs. Given the capabilities of HT-HEDL at handling different numbers of heterogeneous GPUs to improve hypothesis evaluation, as demonstrated by the works on MP-HTHEDL and MP-SPILDL, as well as the experimental results on HT-HEDL, we can clearly demonstrate HT-HEDL's adaptability and hypothesis evaluation capabilities.

\section{Conclusion and future work}
In this work, we focused on accelerating hypothesis evaluation for DL-based ILP learners. We described our hypothesis evaluation engine HT-HEDL, which aggregates the computing power of heterogeneous processors (GPUs with CPUs), to accelerate evaluation performance by orders of magnitude at multiple levels: at the level of individual DL operators, single hypothesis evaluation, and multiple hypotheses evaluation. In individual DL operators, we described CPU-based approaches that combine multi-threading with vector instructions to accelerate computations of DL operators with large speedups. We also described our GPU-based approaches to accelerate computations of DL operators, with which large speedups were achieved.
At the level of a single hypothesis evaluation, we accelerated the computation of covered (positive and negative) examples using multi-threading for CPU-based approaches. For GPU-based approaches, we used warp-based parallel reduction for computing covered examples. Warp-based parallel reduction is regarded as the state of the art for computing reduction operations on GPU hardware. At the level of multiple hypotheses evaluation, we used multi-threading to generate evaluation plans (query plans) for multiple hypotheses in parallel. In addition, we described our multi-device hypothesis evaluation approach, in which HT-HEDL increased hypothesis evaluation throughput by distributing hypotheses to available computing devices (GPUs and a multi-core CPU); all these devices evaluated their assigned hypotheses in parallel. HT-HEDL employs a static scheduling strategy guided by information about the hypothesis evaluation capabilities of each computing device in the system. It estimates the evaluation capabilities of computing devices by evaluating a dummy hypothesis on each computing device in the system against the same knowledge base being used for ILP learning. The execution times for evaluating the probing hypothesis on each computing device, (roughly) reflect the evaluation performance for that given computing device. According to experimental results, HT-HEDL accelerated hypothesis evaluation across the three aforementioned evaluation levels. In fact, HT-HEDL increased hypothesis evaluation throughput up to  44-fold (on a batch of 100 hypotheses) by aggregating the computing power of two GPUs and a single multi-core CPU.
In terms of future research directions, HT-HEDL's work distribution aspects for GPUs can be improved such that each GPU generates its evaluation plans for its assigned hypotheses with minimal to no CPU intervention. This will free the CPU to increase its speedup potential when combined with GPUs (as a single device evaluator). Moreover, as a potential future direction, HT-HEDL can be extended to incorporate FPGA-based accelerators in addition to GPUs and CPUs. In terms of DL, HT-HEDL's hypothesis language can be extended to include more expressive DL constructs (operators).

\section{Acknowledgment}
Special thanks to my PhD supervisor (Dr. Tommy Yuan) from the University of York for his support which helped this research.

\begin{IEEEbiographynophoto}{Eyad Algahtani} received his PhD degree in computer science with focus on scalable machine learning from the university of York, United Kingdom. In his studies, he is working on developing scalable Inductive Logic Programming (ILP) algorithms, capable of constructing human-interpretable Machine Learning (ML) models, from large amount of real-world data; through both HPC (High-performance computing) and non-HPC approaches. He is currently working as an assistant professor in King Saud University, Saudi Arabia. His main research interests focus on scalable and high-performance human-interpretable ML.
\end{IEEEbiographynophoto}

\vfill


\begin{thebibliography}{1}
\bibliographystyle{IEEEtran}
\bibitem{}
Abburu, S. (2012) 'A survey on ontology reasoners and comparison', International Journal of Computer Applications, 57(17), pp. 33-39.

\bibitem{Eyad_2020}
Algahtani, E. (2020) 'A Scalable and Parallel Inductive Learner in Description Logic'. PhD thesis, University of York. 

\bibitem{dl_book_2007}
Baader, F. et al. (2007). The Description Logic Handbook: Theory, Implementation and Applications. Cambridge: Cambridge University Press. 

\bibitem{Blockeel_et_al_2000}
Blockeel, H. et al. (2000) 'Executing query packs in ILP', Proceedings of the 10th International Conference on Inductive Logic Programming (ILP2000), London, July 24-27, pp. 60-77.

\bibitem{}
I. Bratko and S. Muggleton, 'Applications of inductive logic programming', Communications of the ACM, vol. 38, no. 11, pp. 65–70, Nov. 1995.

\bibitem{Chantrapornchai_Choksuchat_2018}
C. Chantrapornchai and C. Choksuchat, 'TripleID-Q: RDF Query Processing Framework Using GPU', IEEE Transactions on Parallel and Distributed Systems, vol. 29, no. 9, pp. 2121–2135, Sep. 2018.

\bibitem{Dean_Ghemawat_2008}
Dean, J. and Ghemawat, S. (2008) 'MapReduce: simplified data processing on large clusters', Communications of the ACM, 51(1), pp. 107-113.

\bibitem{Debnath_et_al_1991}
Debnath, A. K. et al. (1991) 'Structure-activity relationship of mutagenic aromatic and heteroaromatic nitro compounds. Correlation with molecular orbital energies and hydrophobicity', Journal of Medicinal Chemistry, 34(2), pp. 786-797.

\bibitem{Feng_1991}
Feng, C. (1991) 'Inducing temporal fault diagnostic rules from a qualitative model' , Proceedings of the 8th International Conference on Machine Learning (ML'91), Evanston, June 1, pp. 403-406.

\bibitem{}
Flynn, M. J. (1972) 'Some computer organizations and their effectiveness', IEEE Transactions on Computers, C-21(9), pp. 948-960.

\bibitem{Eyad_Kazakov_2018}
Algahtani, E. and Kazakov, D. (2018) 'GPU-Accelerated Hypothesis Cover Set Testing for Learning in Logic', Proceedings of the 28th International Conference on Inductive Logic Programming (ILP2018), Ferrara, September 2, pp. 6-20.

\bibitem{Eyad_Kazakov_2019}
Algahtani, E. and Kazakov, D. (2019) 'CONNER: A Concurrent ILP Learner in Description Logic', Proceedings of the 29th International Conference on Inductive Logic Programming (ILP2019), Plovdiv, September 3, pp. 1-15.

\bibitem{Fonseca_et_al_2009}
Fonseca, N. A. et al. (2009) 'Parallel ILP for distributed-memory architectures', Machine Learning, 74(3), pp. 257-279.

\bibitem{Konstantopoulos_2007}
Konstantopoulos, S. (2007) 'A data-parallel version of Aleph', arXiv, 0708.1527. Available at: https://arxiv.org/pdf/0708.1527.pdf (Accessed 30 Dec 2023).

\bibitem{Martinez_Angeles_et_al_2016}
Martínez-Angeles, C. A. et al. (2016) 'Relational learning with GPUs: accelerating rule coverage', International Journal of Parallel Programming, 44(3), pp. 663-685.

\bibitem{Meissner_2009}
Meissner, A. (2009) 'A simple parallel reasoning system for the ALC description logic', Proceedings of the 1st International Conference on Computational Collective Intelligence (ICCCI 2009), Wroclaw, October 5-7, pp. 413-424.

\bibitem{owl_ref}
Web Ontology Language (2012). Available at: https://www.w3.org/OWL/ (Accessed 22 Mar 2024).

\bibitem{rdf_ref}
Resource Description Framework (2014). Available at: https://www.w3.org/RDF/ (Accessed 22 Mar 2024).

\bibitem{sparql_ref}
SPARQL 1.1 Query Language (2013). Available at: https://www.w3.org/TR/sparql11-query/ (Accessed 22 Mar 2024).

\bibitem{hermit}
Glimm, B. et al. (2014) 'HermiT: an OWL 2 reasoner', Journal of Automated Reasoning, 53(3), pp. 245-269.

\bibitem{pellet}
Sirin, E. et al. (2007) 'Pellet: a practical OWL DL reasoner', Journal of Web Semantics, 5(2), pp. 51-53.

\bibitem{fact++}
Tsarkov, D. and Horrocks, I. (2006) 'FaCT++ description logic reasoner: system description', in Furbach, U. and Shankar, N. (eds.) IJCAR 2006: Automated Reasoning. Berlin: Springer, pp. 292-297.

\bibitem{konclude}
Steigmiller, A., Liebig, T. and Glimm, B. (2014) 'Konclude: system description', Journal of Web Semantics, 27-28, pp. 78-85.

\bibitem{Muggleton_1995}
Muggleton, S. (1995) 'Inverse entailment and progol', New Generation Computing, 13(3), pp. 245-286.

\bibitem{cuda_toolkit}
Nvidia CUDA Toolkit. Available at: https://developer.nvidia.com/cuda-toolkit (Accessed 20 Mar 2024).

\bibitem{intel_sse}
Intel Intrinsics Guide (2023). Available at: https://www.intel.com/content/\\www/us/en/docs/intrinsics-guide/index.html\#techs=SSE\_ALL (Accessed 20 Mar 2024).

\bibitem{openmp}
OpenMP. Available at: https://www.openmp.org/ (Accessed 20 Mar 2024).

\bibitem{OpenMPI_2020}
OpenMPI (2020). Available at: https://www.open-mpi.org/ (Accessed 30 Dec 2023).

\bibitem{gpu_rdf_opt}
P. Makpisit and C. Chantrapornchai, "SPARQL Query Optimizations for GPU RDF Stores", ECTI-CIT Transactions, vol. 17, no. 2, pp. 235–244, Jun. 2023.

\bibitem{gpu_el_reasoner}
Kharma, M. (2017). GRUEL: An EL Reasoner Using General Purpose Computing on a Graphical Processing Unit. Masters thesis, Concordia University. 

\bibitem{gpu_rdf}
P. Makpaisit and C. Chantrapornchai, 'VEDAS: an efficient GPU alternative for store and query of large RDF data sets', Journal of Big Data, vol. 8, no. 1, p. 125, Sep. 2021.

\bibitem{gpu_sparql}
Pang, J. (2020). TriAG:Answering SPARQL Queries Accelerated by GPU. Proceedings of 2020 the 10th International Workshop on Computer Science and Engineering (WCSE2020), Shanghai, 19-21 June, pp. 300-306.

\bibitem{Paes_et_al_2006}
Paes, A. et al. (2006) 'ILP Through Propositionalization and Stochastic k-Term DNF Learning', Proceedings of the 16th International Conference on Inductive Logic Programming (ILP2006),  Santiago de Compostela, August 24-27, pp. 379-393.

\bibitem{Qomariyah_Kazakov_2017}
Qomariyah, N. N. and Kazakov, D. (2017) 'Learning from ordinal data with inductive logic programming in description logic', Proceedings of the 27th International Conference on Inductive Logic Programming (ILP2017), Orléans, September 4, pp. 38-50.

\bibitem{Srinivasan_1999}
Srinivasan, A. (1999) 'A study of two sampling methods for analysing large datasets with ILP', Data Mining and Knowledge Discovery, 3(1), pp. 95-123.

\bibitem{Srinivasan_2007}
Srinivasan, A. (2007) The Aleph manual. Available at: \\https://www.cs.ox.ac.uk/activities/machlearn/Aleph/aleph.html (Accessed 30 Dec 2023).

\bibitem{parting_owl}
S. Priya, Y. Guo, M. Spear and J. Heflin, "Partitioning OWL Knowledge Bases for Parallel Reasoning," 2014 IEEE International Conference on Semantic Computing, Newport Beach, CA, USA, 2014, pp. 108-115

\bibitem{rdf_sq}
J. Urbani and C. Jacobs, 'RDF-SQ: Mixing Parallel and Sequential Computation for Top-Down OWL RL Inference', in Graph Structures for Knowledge Representation and Reasoning, 2015, pp. 125–138.

\bibitem{owl_dl_framework}
Z. Quan and V. Haarslev, "A Framework for Parallelizing OWL Classification in Description Logic Reasoners," arXiv.org, 2019. https://arxiv.org/abs/1906.07749 (accessed Sep. 12, 2024).

\bibitem{par_logic_prog}
A. Dovier, A. Formisano, G. Gupta, M. V. Hermenegildo, E. Pontelli, and R. Rocha, "Parallel Logic Programming: A Sequel," arXiv.org, 2021. https://arxiv.org/abs/2111.11218 (accessed Sep. 12, 2024).

\bibitem{amiri_simd}
H. Amiri, A. Shahbahrami, A. Pohl, and B. Juurlink, "Performance evaluation of implicit and explicit SIMDization," Microprocessors and Microsystems, vol. 63, pp. 158–168, 2018.

\bibitem{maleki_simd}
S. Maleki, Y. Gao, M. J. Garzaran et al., "An evaluation of vectorizing compilers," in Proceedings of the 2011 International Conference on Parallel Architectures and Compilation Techniques (PACT), pp. 372–382, TX, USA, October 2011.

\bibitem{feng_simd}
Feng, J., He, Y., Tao, Q., "Evaluation of Compilers' Capability of Automatic Vectorization Based on Source Code Analysis," Scientific Programming, 2021. https://doi.org/10.1155/2021/3264624

\bibitem{mp_hthedl}
E. Algahtani, "MP-HTHEDL: A Massively Parallel Hypothesis Evaluation Engine in Description Logic," IEEE Access, vol. 12, pp. 89113-89123, 2024, doi: 10.1109/ACCESS.2024.3418815.

\bibitem{mp_spildl}
E. Algahtani, "MP-SPILDL: A Massively Parallel Inductive Logic Learner in Description Logic," in IEEE Access, doi: 10.1109/ACCESS.2024.3458814.

\bibitem{Srinivasan_Faruquie_Joshi_2010}
Srinivasan, A., Faruquie, T. A. and Joshi, S. (2010) 'Exact data parallel computation for very large ILP datasets', Proceedings of the 20th International Conference on Inductive Logic Programming (ILP2010), Florence.

\bibitem{raspberry_pi}
Raspberry Pi 4. Available at: https://www.raspberrypi.org/products/raspberry-pi-4-model-b/ (Accessed 20 May 2023).

\bibitem{eyad_fpga}
E. Algahtani, "A hardware approach for accelerating inductive learning in description logic," ACM Trans. Embedded Comput. Syst., May 2024. https://dl.acm.org/doi/10.1145/3665277.

\bibitem{Zeng_et_al_2014}
Zeng, Q., Patel, J. M. and Page, D. (2014) 'QuickFOIL: scalable inductive logic programming', Proceedings of the VLDB Endowment (PVLDB), 8(3), pp. 197-208. 

\bibitem{ilp_fpga}
A. Fidjeland, W. Luk and S. Muggleton (2002) 'Scalable acceleration of inductive logic programs', Proceedings of the IEEE International Conference on Field-Programmable Technology, Hong Kong, China, 2002, pp. 252-259.
\end{thebibliography}
\end{document}